\documentclass[times,twocolumn]{elsarticle}
\newif\ifanonymous
\newif\ifshowmarkup
\ifanonymous
\newcommand{\ANONYMIZE}[1]{\textit{(omitted for blind reviewing)}}
\else
\newcommand{\ANONYMIZE}[1]{#1}
\fi

\usepackage[table,xcdraw]{xcolor}

\usepackage{color}

\ifshowmarkup
\usepackage[commandnameprefix=always,draft,defaultcolor=red,addedmarkup=colored,deletedmarkup=sout]{changes}
\else
\usepackage[final]{changes}
\fi

\ifshowmarkup
\newenvironment{REVISIONADDENV}{\color{red}}{\color{black}}
\newcommand{\REVISIONADD}[1]{\chadded[]{#1}}
\newcommand{\REVISIONDEL}[1]{\chdeleted[]{#1}}
\newcommand{\REVISIONREP}[2]{\chreplaced[]{#1}{#2}}
\newcommand{\REVISIONCOMMENT}[1]{\chcomment[]{#1}}
\newcommand{\REVISIONIMG}[1]{{\setlength{\fboxsep}{0pt}\setlength{\fboxrule}{2pt}\fcolorbox{red}{yellow}{#1}}}
\else
\newenvironment{REVISIONADDENV}{}{}
\newcommand{\REVISIONADD}[1]{#1}
\newcommand{\REVISIONDEL}[1]{}
\newcommand{\REVISIONREP}[2]{#1}
\newcommand{\REVISIONCOMMENT}[1]{}
\newcommand{\REVISIONIMG}[1]{#1}
\fi

\usepackage{amssymb}
\usepackage{amsmath}
\usepackage{cag}
\usepackage{framed,multirow}

\usepackage{hyperref}
\hypersetup{
    colorlinks=true,
    linkcolor=blue,
    urlcolor=blue,
}

\usepackage{caption}
\usepackage{subcaption}
\begin{document}

\begin{frontmatter}



%%%%%%%%%%%%%%%%%%%%%%%%%%%%%%%%%

\journal{Computers \& Graphics}

\title{Fast and accurate neural reflectance transformation imaging through knowledge distillation}


\ifanonymous
\else
\author[univr]{Tinsae G. Dulecha} %% Author name
\author[univr]{Leonardo Righetto} %% Author name
\author[crs4]{Ruggero Pintus} %% Author name
\author[crs4]{Enrico Gobbetti} %% Author name
\author[univr]{Andrea Giachetti} %% Author name
\affiliation[univr]{organization={University of Verona},%Department and Organization
            %addressline={}, 
            city={Verona},
            %postcode={}, 
            %state={},
            country={Italy}}
\affiliation[crs4]{organization={CRS4},%Department and Organization
            %addressline={}, 
            city={Cagliari},
            %postcode={}, 
            %state={},
            country={Italy}}
\fi

%% Abstract
\begin{abstract}
Reflectance Transformation Imaging (RTI) is very popular for its ability to visually analyze surfaces by enhancing surface details through interactive relighting, starting from only a few tens of photographs taken with a fixed camera and variable illumination. Traditional methods like Polynomial Texture Maps (PTM) and Hemispherical Harmonics (HSH) are compact and fast, but struggle to accurately capture complex reflectance fields using few per-pixel coefficients and fixed bases, leading to artifacts, especially in highly reflective or shadowed areas. The NeuralRTI approach, which exploits a neural autoencoder to learn a compact function that better approximates the local reflectance as a function of light directions, has been shown to produce superior quality at comparable storage cost. However, as it performs interactive relighting with custom decoder networks with many parameters, the rendering step is computationally expensive and not feasible at full resolution for large images on limited hardware. \REVISIONADD{Earlier attempts to reduce costs by directly training smaller networks have failed to produce valid results.} For this reason, we propose to reduce its computational cost through a novel solution based on Knowledge Distillation (\REVISIONREP{DisK}{DISK}-NeuralRTI). Starting from a teacher network that can be one of the original Neural RTI methods or a more complex solution, \REVISIONREP{DisK}{DISK}-NeuralRTI can create a student architecture with a simplified decoder network that preserves image quality and has computational cost compatible with real-time web-based visualization of large surfaces. Experimental results show that we can obtain a student prediction that is on par or more accurate than the existing NeuralRTI solutions with up to 80\% parameter reduction. Using a novel benchmark of high-resolution Multi-Light image collections (RealRTIHR), we also tested the usability of a web-based visualization tool based on our simplified decoder for realistic surface inspection tasks. The results show that the solution reaches interactive frame rates without the necessity of using progressive rendering with image quality loss.

\end{abstract}

\begin{graphicalabstract}
\end{graphicalabstract}

\begin{highlights}
\item Research highlight 1
\item Research highlight 2
\end{highlights}

%% Keywords
\begin{keyword}



\end{keyword}

\end{frontmatter}


\section{Introduction}
\label{sec:intro}

Reflectance Transformation Imaging (RTI) is a computational photography technique widely used, especially in the Cultural Heritage (CH) domain, to interactively \REVISIONREP{inspect a surface by varying its illumination to reveal details}{control the illumination over an investigated surface}. \REVISIONREP{RTI techniques create a relightable image encoding capable of producing pixel colors given a light configuration -- in the most typical case, light intensity and direction. RTI encodings of objects are generated from captured data by analyzing how surface appearance changes under varying light conditions using Multi-Light Image Collections (MLICs), which in the most common case consist of a series of photographs taken from a fixed camera position, with each image illuminated by a differently positioned light~\mbox{\cite{Pintus:2019:SMI}}.}{RTI techniques create a relightable image encoding, i.e., a mapping from light configuration to displayed color, from Multi-Light Image Collections (MLICs). The most common approach considers sets of images captured from a stationary perspective with different lighting angles.}

The most popular and widely utilized encodings are Polynomial Texture Mapping (PTM, \cite{Malzbender:2001:PTM}), which is based on fitting the captured pixel value to second-order polynomial functions of the light direction components, and  Hemispherical Harmonics (HSH)~\cite{Gautron:2004:NHB}, exploiting the hemispherical basis defined from the shifted associated Legendre polynomials. These methods are fast to evaluate and very compact, as they represent per-pixel reflectance fields with few scalar components. They are, thus, the de facto standard for storing, transmitting, and interactively relighting MLICs. These methods, however, are low-frequency and often fail to suitably represent the subtle illumination effects generated by the intertwining of complex local geometric and appearance surface features~\cite{Ponchio:2018:CRR}. For these reasons, see~\autoref{sec:related}, a variety of solutions have been proposed to improve their quality. Recently, neural network-based relightable image encodings~\cite{ren2015image, xu2018deep, Dulecha:2020:NRT} have been demonstrated to greatly enhance the effectiveness of traditional techniques, due to their ability to learn interpolable representations. In this context, NeuralRTI~\cite{Dulecha:2020:NRT} has introduced an effective autoencoder-based solution, which has later been optimized for interactive relighting~\cite{Righetto:2023:EIV,Righetto:2024:EUV}. 

NeuralRTI, however, employs a decoder with numerous parameters\REVISIONADD{, requiring many thousands of arithmetic operations per pixel} to generate the relighted images, which hampers performance and makes them less suitable for real-time interactive object exploration, especially with high-resolution acquisitions. Consequently, recent works focused on improving viewer integration and efficiency by manually optimizing the network's layer count, speeding up the decoding with custom shaders, and implementing a level-of-detail management system that supports fine-grained adaptive rendering through dynamic resampling in the latent feature space~\cite{Righetto:2024:EUV}.

The resulting viewer facilitates the interactive neural relighting of large images, but interactive performances are only achievable through progressive rendering in typical setups, making the user experience far from optimal. A reduction of the decoder complexity could solve this issue, but experimental tests~\cite{Dulecha:2020:NRT, Righetto:2023:EIV} have shown that reducing the number and size of the layers, keeping the same training procedure, results in limited performance boosts without quality degradation. This is because, during training, the limited capacity of the small decoder to model complex reflectance functions makes it difficult to minimize the strongly non-linear loss effectively. As a result, the decoder fails to generalize well, leading to blurry or inaccurate outputs. 

Building on previous work on network compression (\autoref{sec:related}), this work introduces a knowledge distillation technique called DisK-NeuralRTI for compressing the NeuralRTI decoder. Knowledge distillation helps by guiding the small decoder \REVISIONADD{(the \emph{student network})} with the outputs of a well-trained, larger model\REVISIONADD{ (the \emph{teacher network})}, simplifying the learning task and enabling better training convergence and performance. As a result, the method enables the production of high-quality relighted images with a limited fraction of the original decoding parameters, making it possible to perform a smooth interactive relighting even in the case of large images and limited computational power. To the best of our knowledge, this is the first work applying this approach to the RTI relighting domain. The results show that the resulting solution outperforms the manual tuning and is highly effective, making the Neural RTI encoding usable in practical settings. 

This paper is the extension of a work published in the proceedings of STAG 2024~\cite{drp24}. 
%In addition to improving the presentation, we introduce significant new material.
While\REVISIONADD{,} in the original paper, we applied the novel training procedure just to perform data reduction on the original NeuralRTI model, \REVISIONREP{this work also evaluates}{we also evaluate} the effect of using different teacher architectures. \REVISIONADD{Furthermore, we improve the evaluation by introducing four additional datasets for testing both relighting quality and rendering efficiency for different types of surfaces and materials.}
Finally, we also evaluate the time needed to train the networks and show how to speed up the training procedure for large images and costly teacher networks.

The rest of the article is organized as follows:  \autoref{sec:related} presents related work on surface relighting with RTI, neural relighting, and network compression; \autoref{sec:method} describes the proposed knowledge distillation solution with improved teachers and a lightweight student network. The validation of the proposed approach on the RealRTI and SynthRTI benchmarks are reported in~\autoref{sec:results}, while~\autoref{sec:results:chmlic} presents the benchmark comprising high resolution Multi-Light Image collections and the results obtained with our approach on it in terms of accuracy and interactive relight performances, as well as tests on strategies for downsampling of input pixels for the speed-up of the training procedure. Finally, \autoref{sec:discussion} summarizes our findings and discusses potential future works.

\section{Related Work}
\label{sec:related}
RTI is routinely used in the CH domain to analyze surface properties~\cite{Pintus:2019:SMI} and is also employed in other domains, such as manufacturing~\cite{nurit2021hd} and quality assessment~\cite{coules2019reflectance}. The goal of RTI is mainly to support interactive relighting, simulating the inspection of a surface with a manually controlled illumination direction. Technically, it relies on a pixel-space encoding of the reflectance behavior of the surface, depending on both shape and material properties, estimated from a Multi-Light image collection.
The three main characteristics that these encodings must possess are compactness, to simplify end-to-end storage and transfer of relightable image data; smooth interpolation/approximation, to provide the illusion of continuous control of light direction; and speed, to support interactive relighting of high-resolution images on high-pixel-count displays without quality degradation. In the following, we briefly summarize the approaches that have been used to achieve these goals using \REVISIONADD{shape and material separation (\mbox{\autoref{sec:related:separate}})}, classical (\autoref{sec:related:classical}\REVISIONADD{), } and neural (\autoref{sec:related:neural}) RTI techniques, before discussing which neural network compression methods proposed in the literature are more appropriate for our use case (\autoref{sec:related:neural-compression}).

%------------------------------------------------------------------------------
\subsection{\REVISIONREP{Shape and material separation}{Classical RTI}}
\label{sec:related:separate}

A first approach to provide a compact and fast encoding of the reflectance field is to separate the shape and material components to generate a physically-based representation of the interaction between the light and the imaged object. In MLIC scenarios, such decoupled representations typically combine per-pixel maps of normals and Spatially-Varying Bidirectional Reflectance Distribution Functions (SV-BRDFs). Recovering this representation from MLIC data via photometric stereo and BRDF fitting is, however, challenging, as single-view, multi-illumination setups capture only a sparse slice of the BRDF, leading to the need for multi-view acquisitions or the use of strong analytical or learned priors~\cite{Guarnera:2016:BRA,Zhang:2021:NNF,Pintus:2023:ELS}. Moreover, and most importantly, while decoupling shape and material can be effective, these methods are difficult to derive from commonly available sampled data and do not generalize easily across diverse object classes and material behaviors, for instance, semitransparent and multilayered objects~\cite{Guarnera:2016:BRA,Pintus:2023:ELS}. As a result, \REVISIONADD{} relighting approaches instead directly approximate the reflectance by directly mapping lighting parameters (mostly direction) to final observed values, bypassing any explicit separation of shape and material~\cite{Pintus:2019:SMI,Zhang:2014:ERI}.

%------------------------------------------------------------------------------
\subsection{\REVISIONADD{Classical RTI}}
\label{sec:related:classical}

Polynomial Texture Mapping (PTM,~\cite{Malzbender:2001:PTM,Zhang:2014:ERI}) and Hemispherical Harmonics (HSH,~\cite{Gautron:2004:NHB}) are the earlier and still most widely used compact, low-complexity reflectance field encodings proposed for relighting. They fit simple parametric functions of the light direction components to the local MLIC pixel values. 
They \REVISIONREP{can render}{are highly efficient for rendering} relighted images given a novel input direction \REVISIONREP{using}{as they require} only a few arithmetic operations per pixel, but \REVISIONREP{can only reproduce}{are accurate only in the reproduction of} relatively low-frequency, smooth behaviors~\cite{Pintus:2019:SMI}. \REVISIONADD{While the method achieves good results, especially for reflective surfaces, the behavior is similar to PTM and HSH when using a number of modes compatible with interactive reconstruction~\mbox{\cite{pitard2017discrete}}.} Low-frequency reconstructions were improved by separately modeling matte behaviors and high-frequency ones. In particular, several authors~\cite{drew2012robust,Zhang:2014:ERI,Fornaro:2017:ERG} \REVISIONREP{have proposed using}{proposed to use} PTM or HSH for matte modeling, and \REVISIONDEL{using }a separate detail map to approximate the difference between the matte model and the original images. The storage cost of the detail coefficient map is, however, high. The multi-scale structure of the reflectance field was also harnessed by introducing Discrete Modal Decomposition (DMD)~\cite{pitard2017discrete}. \REVISIONDEL{While the method achieves good results, especially for reflective surfaces, the behavior is similar to PTM and HSH when using a number of modes compatible with interactive reconstruction~\mbox{\cite{pitard2017discrete}}.}  In~\cite{Giachetti:2018:NFH}, Radial Basis Function (RBF) interpolation of the original data has been proposed as an alternative idea, but the method requires run-time access to the original image data and cannot provide interactive relighting. It was later combined with Principal Component Analysis (PCA) compression of the image stack and RBF interpolation in light space to improve efficiency at the cost of a slight reduction in quality~\cite{Ponchio:2018:CRR}. 

Thanks to their versatility, compression rate, and decoding speed, PTM, HSH, and/or RBF+PCA are the encodings used in most of the publicly available web-based tools for image-based relighting, e.g., \emph{WebRTIViewer}~\cite{webrtiviewer}, \emph{Pixel+~Viewer}~\cite{ppviewer},  
\emph{Marlie}~\cite{Jaspe:2021:WEA}, \emph{Relight}~\cite{relight,Ponchio:2018:CRR}, and \emph{OpenLIME}~\cite{OpenLime}\REVISIONADD{\mbox{\cite{Ponchio:2025:OOF}}}. In this work, we aim to provide a plug-in solution based on neural compression \REVISIONADD{(see~\mbox{\autoref{sec:related:neural}}), to improve quality, especially for high-frequency reflectance components, at storage and timing costs similar to standard solutions}. 

%------------------------------------------------------------------------------
\subsection{Neural-based RTI} 
\label{sec:related:neural}

In recent years, neural networks have proven effective for compression, nonlinear approximation, and interpolation of large datasets, and these properties have found applications in rendering tasks~\cite{Ren:2015:IBR,Xu:2018:DIR,Tewari:2020:SAN}. Specifically, the NeuralRTI method~\cite{Dulecha:2020:NRT} was introduced as a direct alternative to traditional RTI representations. The method exploits a fully connected asymmetric autoencoder to represent the original per-pixel information with a low-dimensional vector. The network is trained end-to-end to accurately reproduce the training pixel values. After training, the encoder is discarded, and the resulting decoder can be used to relight the image using a novel light direction combined with stored latent features. Pistellato et al.~\cite{Pistellato:2023:ORT} proposed a modified approach, training the decoder with PCA-compressed data to deal with a large number of input images. Due to non-linearity, the method improves over the other classical RTI solutions in reproducing complex light scattering properties and shadows, but at the cost of a computational complexity of the decoding at least two orders of magnitude higher. This fact impacts the interactivity of practical applications, particularly in the Cultural Heritage domain, where the classical techniques are still the most popular ones. 

To address this issue, Righetto et al.~\cite{Righetto:2023:EIV} introduced a modified version of the original Neural RTI. Through a series of experiments, they manually reduced the complexity while maintaining relighting quality. However, this reduction was insufficient to ensure interactive relighting for high-resolution images in the case of limited computational power or screens with large pixel counts. Simplified decoders were tested but did not converge to good-quality representations due to the complexity of the error landscape. For this reason, they later improved interactivity by performing the decoding directly within pixel shaders and by using an adaptive multi-resolution renderer to meet \REVISIONADD{rendering} frequency requirements\REVISIONADD{ during interactive exploration}~\cite{Righetto:2024:EUV}. This solution, however, does not provide full-quality images during relighting or when panning over high-resolution images on high-pixel-count displays~\REVISIONADD{, since the presented image is computed at a lower scale and upscaled at presentation time}. In this work, we aim to reduce the decoding complexity by exploiting automated solutions to build reduced networks.

%------------------------------------------------------------------------------
\subsection{Neural Network Compression}
\label{sec:related:neural-compression}

A reasonable idea to speed up decoding is to exploit automated network compression techniques recently proposed in the literature~\REVISIONADD{ to reduce per-pixel computing costs}. These techniques are categorized into four major categories: parameter pruning, low-rank factorization, network quantization, and knowledge distillation. 

Parameter pruning methods~\cite{han2015deep,guo2016dynamic,yao2017deepiot} concentrate on locating and eliminating redundant or non-essential parameters from models. \REVISIONREP{These}{Although these} approaches can achieve significant compression rates and reduce the number of arithmetic operations. However, they often require transforming fully-connected layers into sparsely-connected configurations. This change can complicate the decoding process in GPU shaders and may hinder performance, particularly for small networks like NeuralRTI. 

Low-rank factorization methods~\cite{tai2015convolutional,bhattacharya2016sparsification} employ matrix and tensor decomposition techniques to pinpoint the essential parameters in convolutional neural networks (CNNs). However, these strategies generally produce remarkable outcomes primarily for moderately large to very large networks, which are many orders of magnitude larger than the NeuralRTI decoder. 

Network quantization techniques~\cite{gong2014compressing,courbariaux2015binaryconnect,hubara2016binarized} lessen the number of bits needed to represent each weight, resulting in a compressed network. This size reduction \REVISIONREP{may also accelerate}{also accelerates} processing by enhancing cache efficiency, but, on its own, it can only achieve a moderate speed-up when limited to data types supported by GPUs. 

Lastly, knowledge distillation~\cite{hinton2015distilling} focuses on training a more compact (student) model to imitate the behavior of a larger (teacher) model, transferring knowledge from the expansive network to the smaller one, while preserving prediction accuracy. Consequently, the student model learns to replicate the predictions of the teacher. Initially intended for classification tasks, this method has also been utilized in regression scenarios (e.g.,~\cite{takamoto2020efficient,saputra2019distilling}).  This approach is effective for regressing complex nonlinear responses because the teacher model captures high-level patterns and smooth approximations of the target function that are difficult for a small model to learn directly. By mimicking the teacher’s outputs, the student learns a simplified version of the complex response surface, making training more stable and enabling better generalization despite its limited capacity. 

In our work, we use knowledge distillation to create a lightweight RTI decoder trained to imitate the original RTI by following the behavior of a more complex teacher network. To the best of our knowledge, we are the first to implement knowledge distillation in the context of neural relighting based on MLICs.

%-------------------------------------------------
\begin{figure}[!ht]
     \centering
     \begin{subfigure}[b]{0.98\linewidth}
         \centering
         \REVISIONIMG{\includegraphics[width=\textwidth]{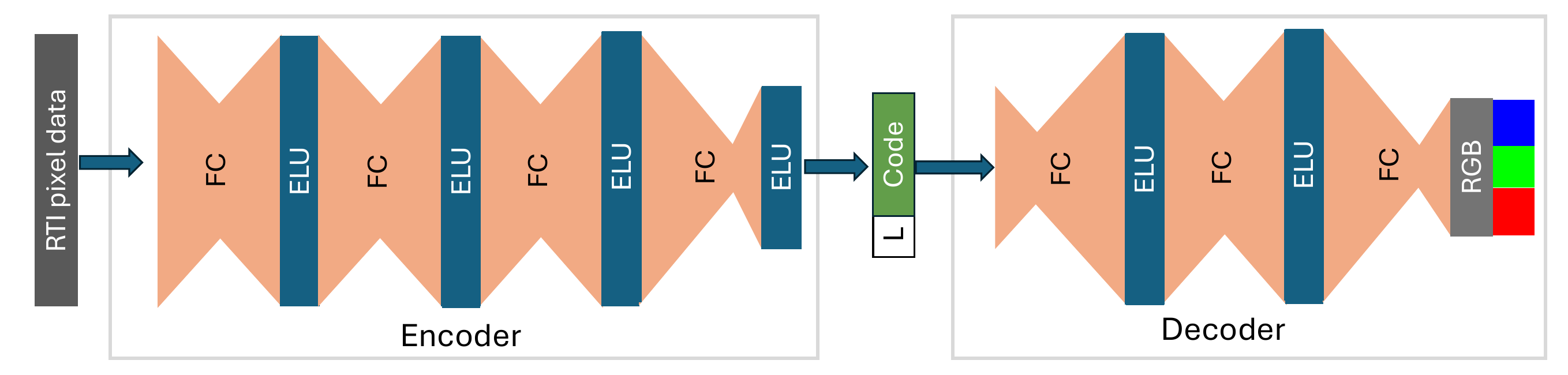}}
         \caption{NeuralRTI scheme. The encoder has three hidden layers\REVISIONADD{,} and the decoder has two hidden layers.  The encoder comprises hidden layers with 150 units each, and the decoder comprises hidden layers with 50 units each. The encoder receives RTI pixel data and produces a 9-dimensional code. The decoder concatenates the code vector with the light direction and outputs a single RGB value. }
         \label{fig:NeuralRTI}  
     \end{subfigure}

          \begin{subfigure}[b]{0.98\linewidth}
         \centering
       \REVISIONIMG{\includegraphics[width=\textwidth]{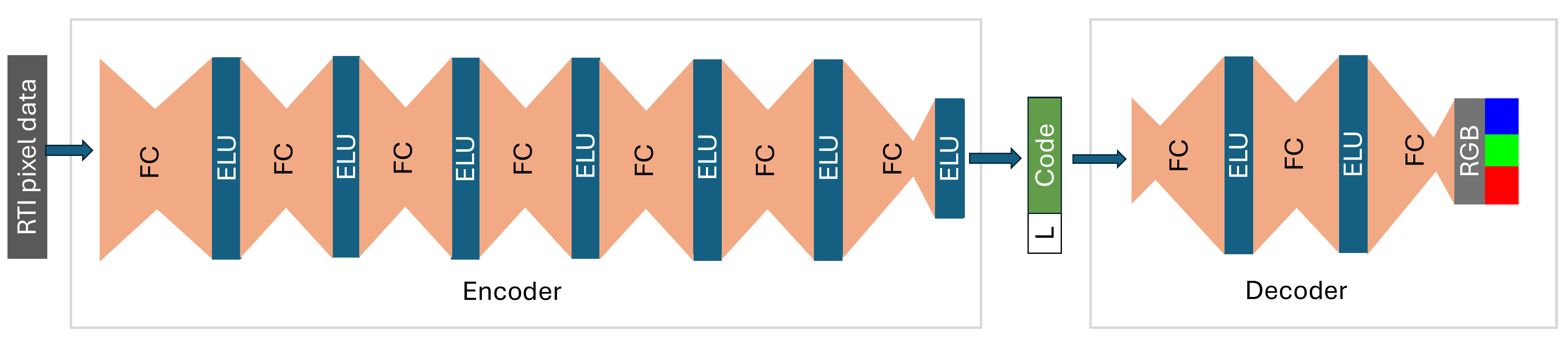}}         \caption{Improved teacher network. The encoder consists of six hidden layers instead of three, like the original architecture (a).}
         \label{fig:Improved}  
     \end{subfigure}
     
     \begin{subfigure}[b]{0.98\linewidth}
         \centering
         \REVISIONIMG{\includegraphics[width=\textwidth]{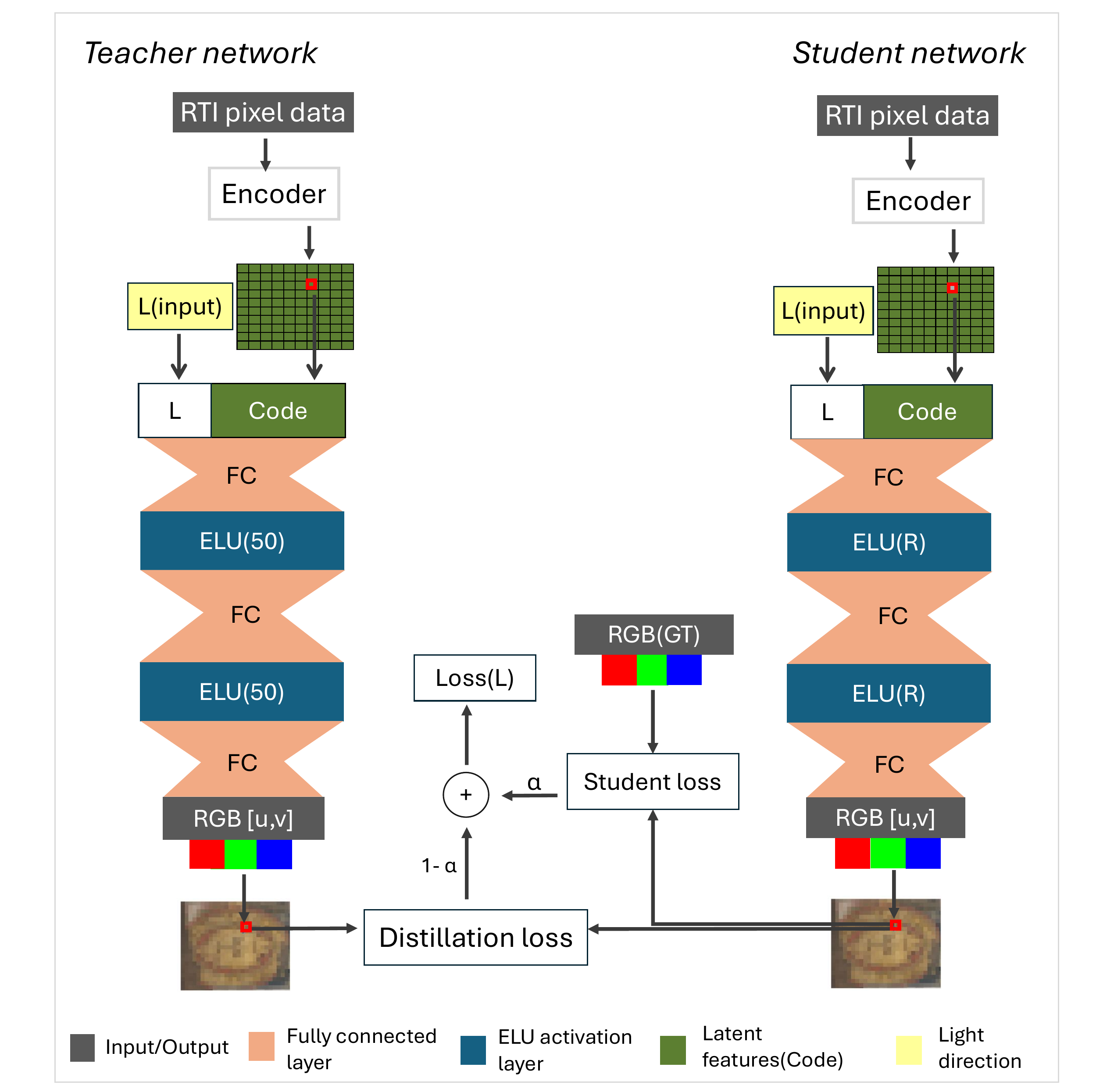}}
         \caption{DisK-NeuralRTI. The encoder has the same architecture for student and teacher networks. The student network decoder contains two layers, each with an R number of units. We tested it with R values of $10$ and $20$.}
         \label{fig:DiskNeuralRTI}
     \end{subfigure}
  \caption{Network architecture for original NeuralRTI (top) and DisK-NeuralRTI (bottom). }\label{fig:all_networks}
\end{figure}
%-------------------------------------------------

\section{DisK-NeuralRTI}
\label{sec:method}

\REVISIONADD{In this work, we introduce a knowledge distillation technique to improve the efficiency of neural relighting. While the method is generally applicable to reduce the size and complexity of general neural network solutions, we specialize it here for NeuralRTI~\cite{Righetto:2023:EIV,Righetto:2024:EUV}. In the following, we will first provide  Background on the original NeuralRTI network (\autoref{sec:method-original-neuralrti}). We will then discuss the structure and parameterization of the student (\autoref{sec:method-student}) and teacher (\autoref{sec:method-teacher}), as well as the training strategy for performing the optimization (\autoref{sec:method-training}).}

%------------------------------------------------------------------------------
\subsection{\REVISIONADD{Background: The NeuralRTI network}}
\label{sec:method-original-neuralrti}

\REVISIONADD{The NeuralRTI network, as introduced in the work by Righetto et al.~\cite{Righetto:2023:EIV,Righetto:2024:EUV}, is illustrated in~\autoref{fig:NeuralRTI}.}
\REVISIONDEL{Our approach is based on the NeuralRTI network illustrated in~\mbox{\autoref{fig:NeuralRTI}}, that has been used in the work by Righetto et al.~\mbox{\cite{Righetto:2023:EIV,Righetto:2024:EUV}}.}

In this model, the encoder comprises four layers, each with an Exponential Linear Unit (ELU) activation function. It processes per-pixel RTI data, which includes pixel values for the sampled lighting directions, and compresses this data into nine-dimensional latent space features. The decoder network, shown on the right in~\autoref{fig:NeuralRTI}, includes two hidden layers, each with 50 units. It takes the concatenation of the pixel encoding and a 2D vector representing the light direction as input. The decoder's output is the predicted RGB pixel value illuminated from the specified light direction. 
The network is trained end-to-end on the pixels of the original MLIC (all, or a subset), minimizing the mean squared error between the predicted pixel values and the ground truth ones across the specified light directions. 

Once the training phase is complete, the encoder is used to generate the final latent features for each pixel. Latent feature values are stored as relightable image data, and the encoder is then discarded. To compute relighted images, the learned decoder is then used to process the latent features, combined with the light directions interactively set by the user, to produce the final output.

The goal of this work is to use knowledge distillation to train a simple version of this network in a way that is not achievable directly by training it on the raw MLIC data.

%------------------------------------------------------------------------------
\subsection{Student network}
\label{sec:method-student}
The student network (\autoref{fig:DiskNeuralRTI}, right) is designed to simplify only the decoder component, since the encoder size does not impact interactive relighting. We design it to be a version of the NeuralRTI model with the same encoder but a smaller decoder.

In the NeuralRTI model, the total number of decoder weights ($W$) and biases ($B$) is given by $W = (K+2) \times N + N \times N + N \times 3$ and $B = N + N + 3$. The $K$ latent code values are stored per pixel, and setting $K=9$ leads to a compression rate similar to standard PTM. Decoding complexity is mostly due to the number of decoder weights and biases. With decoder layers of size $N=50$, as in the original network, the number of weights and related multiplications is $W=3200$, and the number of biases and related additions is $B=103$. 

To achieve a sensible speed-up, we want to decrease the number of units in each hidden decoder layer from the original $50$ to significantly smaller values, aiming to reduce the total number of parameters in the decoder while ensuring interactive relighting for large images, even with limited hardware resources. 

Setting the decoder layer size to $N=20$ reduces the number of weights and multiplications to $W=680$ and the number of biases and sums to $43$, achieving an $80\%$ reduction in computation and memory fetches. A layer size of $N=10$, leads, instead, to $W=240$ weights and multiplications, and $23$ biases and sum\REVISIONADD{s}, giving a $92\%$ reduction in computational cost and memory pressure. The theoretical speed-up is thus very significant: between $\approx 5 \times$ and $\approx 10 \times$ for these configurations.
%Both cases were tested, and these adjustments effectively maintained smooth interactive relighting during our evaluations while achieving relighting accuracy comparable to the original NeuralRTI (see \autoref{sec:results}). 

%------------------------------------------------------------------------------
\subsection{Teacher network}
\label{sec:method-teacher}

The teacher network must first be able to learn a representation from the raw data directly, and then, once trained on the raw data, serve as guidance to the student network during the distillation phase. Since the original NeuralRTI autoencoder was proven to be able to perform relighting by learning from MLICs, we used it as a teacher in our original conference publication~\cite{drp24}. Since the efficient relighting is provided by the student network decoder, we are not constrained to use the original, manually optimized, NeuralRTI architecture as a teacher.

We tested several designs for a more complex teacher network, changing the number of encoding layers, increasing the layers' size, adding skip connections, and also trying to add the light direction information in the input pixel information, concatenated to the color data.
The best results in preliminary testing on a subset of the MLIC considered were obtained with the solution shown in~\autoref{fig:Improved}, as also illustrated in~\autoref{sec:results}.

%------------------------------------------------------------------------------
\subsection{Training}
\label{sec:method-training}

Irrespective of the teacher used, we train the student network on all pixels of the original MLIC (or a subset of it) by minimizing the following loss function: 
\begin{equation}
\label{loss}
    L= \frac{1}{n} \sum_{k=1}^{n} \alpha  {|| P_s - P_{gt}||}_k ^ 2   + (1 -\alpha)  {|| P_s - P_{t}||}_k ^ 2  
\end{equation}

This function is a weighted combination of two components: the student loss, which measures the difference between the student’s predictions ($P_s$) and the ground truth pixel values ($P_{gt}$), and the distillation loss, which measures the difference between the teacher’s predictions ($P_t$) and the student’s predictions ($P_s$). The parameter $\alpha$ determines the weight of each loss component. The distillation loss captures instead the discrepancy between the student and teacher models. Minimizing this loss during training enhances the student model’s ability to replicate the teacher’s predictions accurately. 

The basic idea behind the approach is that training a very compact model through distillation should be more effective than training it directly on the original data. Fitting original data with the larger teacher network is easier than fitting it with the smaller student network, thanks to the larger number of parameters. At the same time, the teacher model’s outputs are typically smoother/less noisy and may contain richer information than the exact regression target values coming from the original images. During distillation, the teacher model can thus provide hints about the underlying distribution of the data, which can guide the student model to learn more effectively to fit the original data and generalize better. 
%The experiments are implemented using PyTorch on four NVIDIA Ampere A100 GPUs, 64GB each. 

If the input MLIC is composed of multiple high-resolution images, the cost of training may be very high, especially when using a more complex teacher network.
However, as many of the pixels, especially in the close neighborhood, often contain very similar information, there is no need to use all the pixels to train the model on all of them. We therefore experimented with learning DisK\REVISIONADD{-}NeuralRTI from only a subset of them. Currently, we subsample the input images before training using uniform random sampling to avoid introducing biases due to alignment with image features that could appear using a regular subsampling pattern. \autoref{sec:results} illustrates the effects of training on reduced data in terms of speed and quality of achieved results.

\section{Results and Evaluation}
%datasets: SynthRTI and RealRTI}
\label{sec:results}
A first set of experimental tests was aimed at assessing the benefits of the new training approach, determining a reasonable layer size for the simplified decoder, and showing the quality improvements obtained with the new teacher network. These tests were performed on existing low-resolution relighting benchmarks proposed in previous works~\cite{Dulecha:2020:NRT}, namely SynthRTI and RealRTI.

SynthRTI~\cite{SynthRTI} is a collection of $51$ synthetic MLICs rendered with the Blender Cycles engine. It is divided into two parts: SingleMaterial, featuring $24$ collections created from three surfaces with $8$ different materials applied,
%\FIXME{(see Figure X)},
and MultiMaterial, including $27$ collections created with the same 3 surfaces painted with $9$ material combinations each. The simulated materials make it possible to evaluate how well the relighting techniques handle a wide range of reflective behaviors.
%\FIXME{(see Figure X)}. 
Each collection is split into two sets of images, corresponding to two separate groups of light directions. The first set, called \textit{Dome}, corresponds to a multi-ring light dome configuration with $49$ directional lights arranged in concentric rings in the $l_x, l_y$ plane at $5$ different elevation angles ($10, 30, 50, 70, 90$ degrees). The second, called \textit{Test}, corresponds to other $20$ light directions at $4$ intermediate elevation angles ($20, 40, 60, 80$ degrees). We used the \textit{Dome} subset to train the networks and the \textit{Test} set to evaluate the quality of the relighted images.

RealRTI~\cite{RealRTI} includes 12 multi-light image collections derived from real CH-related acquisitions, but cropped and resized to allow fast processing/evaluation.
MLICs were acquired with both light domes and handheld RTI protocols \cite{Pintus:2019:SMI} and feature surfaces with different complexity in shape and material. In the original paper~\cite{Dulecha:2020:NRT}, the testing protocol for validating the relighting was based on a leave-one-out training and testing methodology. We, instead, decided to use the same testing protocol of SynthRTI, removing 5 test images at different elevations for each collection (used then as the test set) and training the relightable images on the remaining ones. This makes the test faster but similarly informative and challenging.

\REVISIONADD{For all the benchmarks, in this extended paper, we also evaluated additional metrics, namely LPIPS \cite{zhang2018perceptual} and DeltaE \cite{mokrzycki2011colour}, to assess the ability of the method to preserve perceptual similarity and chromatic information.}

%We evaluated our DisK-NeuralRTI compression using two approaches. First, we assessed the quality of relighted images at different compression levels on the same benchmarks used to showcase NeuralRTI's advantages \cite{Dulecha:2020:NRT}, namely SynthRTI and RealRTI. SynthRTI includes synthetic multi-light image collections with various shapes and material combinations, while RealRTI contains real captures of diverse surfaces. Additionally, we tested the compressed network's capability to deliver real-time relighting for real-world use cases stemming from the cultural heritage area, where we applied the method to novel high-resolution surface captures. On these new datasets, we demonstrated the increased rendering speed achieved with DisK-NeuralRTI compression using the OpenLIME visualization framework.

\begin{figure}[!ht]
\centering
  \begin{subfigure}[b]{1\linewidth}
    \REVISIONIMG{\includegraphics[width=1\linewidth]{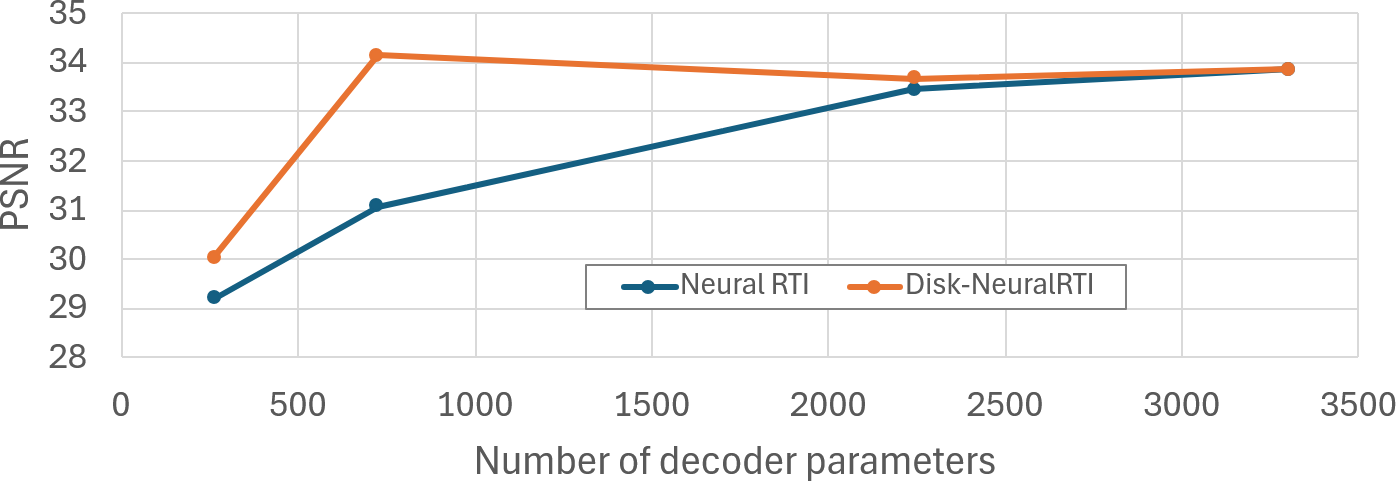}}
    \caption{}
    \label{fig:line_synthrtiS}
\end{subfigure}
  \begin{subfigure}[b]{1\linewidth}
    \REVISIONIMG{\includegraphics[width=1\linewidth]{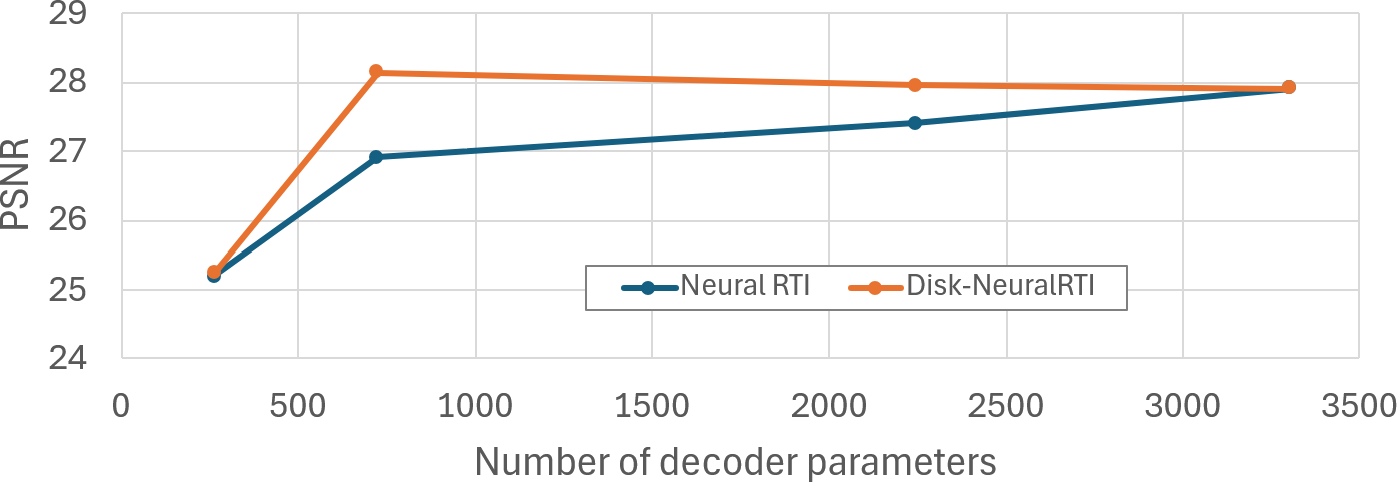}}
    \caption{}
    \label{fig:line_synthrtiM}
\end{subfigure}
  \begin{subfigure}[b]{1\linewidth}
    \REVISIONIMG{\includegraphics[width=1\linewidth]{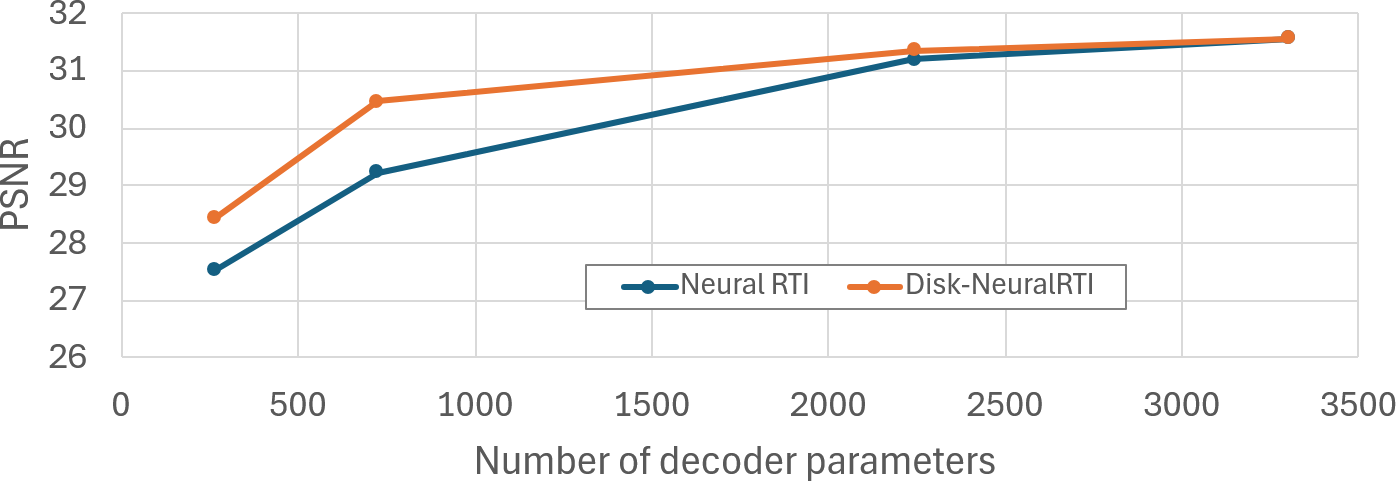}}
    \caption{}
\label{fig:real_rti} 
\end{subfigure}
 \caption{Line charts showing the relighting quality (PSNR) as a function of the number of decoder parameters for the SynthRTI Multi-Material benchmark (a), the SynthRTI Multi-Material benchmark (b), the RealRTI benchmark (c).  Training with DisK\REVISIONADD{-}NeuralRTI results in metrics close to or better than the teacher for the 723 parameters version.}
\end{figure}

%---------------------------------------------------------------------------
\subsection{Setup and parameter tuning}
\label{sec:results:setup}
For all the NeuralRTI training procedures, we used  $90\%$ of the total (or subsampled) RTI data pixels for training and reserved $10\%$ for validation. We chose the Adam optimization algorithm \cite{kingma2014adam}, with a batch size of $64$, a learning rate of $0.01$, a gradient decay factor of $0.9$, and a squared gradient decay factor of $0.99$. 
For the distillation tasks, we determined with a preliminary test a single value of the parameter $\alpha$ and used it for all the benchmarks. In detail, we
%The selection was made by computing 
evaluated the average relighting quality on five real captures from the ReaRTI dataset, varying the value of $\alpha$ in the range [0.1-0.9]. It was possible to observe that the method is not very sensitive to variations of $\alpha$ in the range [0.1..0.7], and there is a drop in quality only when alpha exceeds 0.8. We therefore set $\alpha=0.6$.

The network implementation leverages the PyTorch library.
%, and the decoder was then converted to an OpenGL shader for real-time display, using the methods described in ~\cite{Righetto:2023:EIV,Righetto:2024:EUV}.
After training, per-pixel latent codes are converted to single bytes using offset/scale mapping and then stored as image byte planes. The decoder parameters (weights and biases) and the metadata are saved in a JSON file.
The JSON header and the images with the latent codes can be used by an interactive web viewer, \REVISIONREP{which}{that} is based on a custom shader that reads the decoder data and executes the decoding code, providing a fast estimation of the imaged surface religthed under novel, arbitrary light directions~\cite{Righetto:2023:EIV,Righetto:2024:EUV}.

%  %

\begin{table*}[ht]
\footnotesize
\centering
\setlength{\tabcolsep}{2pt}
\begin{tabular}{|l|c|c|c|c|c|c|c|c|}
\hline
  \cellcolor[HTML]{EFEFEF}&
  \cellcolor[HTML]{EFEFEF}\begin{tabular}[c]{@{}c@{}}NeuralRTI\\ (50)\end{tabular} & 
  \cellcolor[HTML]{EFEFEF}\begin{tabular}[c]{@{}c@{}}NeuralRTI\\ (20)\end{tabular} & 
    \cellcolor[HTML]{EFEFEF}\begin{tabular}[c]{@{}c@{}}DisK-N\REVISIONADD{eural}RTI\\ (20)\end{tabular} &
  \cellcolor[HTML]{EFEFEF}\begin{tabular}[c]{@{}c@{}}NeuralRTI-IT\\ (50)\end{tabular} &

  \cellcolor[HTML]{EFEFEF}\begin{tabular}[c]{@{}c@{}}DisK-N\REVISIONADD{eural}RTI-IT\\ (20)\end{tabular} &
  \cellcolor[HTML]{EFEFEF}PTM &
  
  \cellcolor[HTML]{EFEFEF}HSH 3ord &
  \cellcolor[HTML]{EFEFEF}PCA/RBF \\ \hline
  \cellcolor[HTML]{EFEFEF}Canvas     & 41.42/0.99 &  39.88/0.99 & 40.89/0.99 & \textbf{46.15/0.99}  & 44.68/0.99 & 29.03/0.98 &  41.24/0.99 & 34.2/0.99  \\ \hline
\cellcolor[HTML]{EFEFEF}Tablet     & 29.13/0.88 & 26.45/0.83 & 30.53/0.90
  & \textbf{31.03/0.91}  & 29.78/0.89  & 23.79/0.81  & 29.92/0.87 & 25.87/0.80 \\ \hline
\cellcolor[HTML]{EFEFEF}Bas-relief & 31.02/0.89 & 26.91/0.83 & 30.97/0.90 & \textbf{34.08/0.94}  & 31.31/0.90 & 24.47/0.81  & 28.82/0.86 & 25.55/0.86 \\ \hline
\rowcolor[HTML]{EFEFEF} 
Average                            & 33.86/0.92  & 31.08/0.88 & 34.13/0.93  &  \textbf{37.09/0.95}& 35.26/0.93 & 25.76/0.87   & 33.33/0.91 & 28.54/0.88 \\ \hline
\end{tabular}
\caption{Average PSNR/SSIM values for the relighting of test images of SynthRTI SingleMaterial collections. DisK-NeuralRTI provides very good results with a per-pixel encoding size of 9 parameters (as PTM and PCA/RBF) and a sufficiently small number of shared decoding parameters per image. With a layer size of 20 elements. it provides better metrics than the teacher networks. Bold figures indicate the best values. Figures in parentheses indicate the network layers' size.}
\label{tab:synthRTISingle}
\end{table*}

\begin{table*}[ht]
\footnotesize
\centering
\setlength{\tabcolsep}{2pt}
\begin{tabular}{|l|c|c|c|c|c|c|c|c|}
\hline
 \cellcolor[HTML]{EFEFEF}&
  \cellcolor[HTML]{EFEFEF}\begin{tabular}[c]{@{}c@{}}NeuralRTI\\ (50)\end{tabular} &
  \cellcolor[HTML]{EFEFEF}\begin{tabular}[c]{@{}c@{}}NeuralRTI\\  (20)\end{tabular} &
  \cellcolor[HTML]{EFEFEF}\begin{tabular}[c]{@{}c@{}}DisK-N\REVISIONADD{eural}RTI\\ (20)\end{tabular} &
    \cellcolor[HTML]{EFEFEF}\begin{tabular}[c]{@{}c@{}}NeuralRTI-IT\\  (50)\end{tabular} &
  \cellcolor[HTML]{EFEFEF}\begin{tabular}[c]{@{}c@{}}DisK-N\REVISIONADD{eural}RTI-IT\\ (20)\end{tabular} &
  \cellcolor[HTML]{EFEFEF}PTM &
  \cellcolor[HTML]{EFEFEF}HSH 3 ord &
  \cellcolor[HTML]{EFEFEF}PCA/RBF \\ \hline
\cellcolor[HTML]{EFEFEF}Canvas     & 33.33/0.96 &31.94/0.95  & 32.63/0.95  & \textbf{35.94/0.97} & 33.21/0.95 & 25.17/0.93  & 30.03/0.95 & 27.95/0.95 \\ \hline
\cellcolor[HTML]{EFEFEF}Tablet     & 24.29/0.77 & 23.58/0.75 & 24.96/0.79 & \textbf{25.57}/0.80  & 24.91/0.79 & 20.56/0.79  & 24.24/\textbf{0.84} & 20.89/0.76 \\ \hline
\cellcolor[HTML]{EFEFEF}Bas-relief & 26.08/0.83 & 25.20/0.80  & 26.83/0.84 & \textbf{27.98/0.86}  & 26.48/0.83 & 22.34/0.76  & 25.10/0.79 & 21.54/0.81 \\ \hline
\rowcolor[HTML]{EFEFEF} 
Average                            & 27.90/0.85 & 26.91/0.83 & 28.14/0.86  &  \textbf{29.83/0.88}  & 28.20/0.86 & 22.69/0.83  & 26.46/0.86 & 23.46/0.84 \\ \hline
\end{tabular}
\caption{Average PSNR/SSIM values for the relighting of test images of SynthRTI MultiMaterial collections. The 20-elements layer compression achieves better results than the teacher network. Figures in parentheses indicate the network layers' size.}
\label{tab:synthRTIMulti}
\end{table*}

% \usepackage{graphicx}
\begin{table*}[ht]
\footnotesize
\centering
\setlength{\tabcolsep}{2pt}
\begin{tabular}{|l|c|c|c|c|c|c|c|c|}
\hline
 \cellcolor[HTML]{EFEFEF}&
  \cellcolor[HTML]{EFEFEF}\begin{tabular}[c]{@{}c@{}}NeuralRTI\\ (50)\end{tabular} &
    \cellcolor[HTML]{EFEFEF}\begin{tabular}[c]{@{}c@{}}NeuralRTI\\ (20)\end{tabular} &
      \cellcolor[HTML]{EFEFEF}\begin{tabular}[c]{@{}c@{}}DisK-N\REVISIONADD{eural}RTI\\ (20)\end{tabular} &
  \cellcolor[HTML]{EFEFEF}\begin{tabular}[c]{@{}c@{}}NeuralRTI-IT\\  (50)\end{tabular} &
  \cellcolor[HTML]{EFEFEF}\begin{tabular}[c]{@{}c@{}}DisK-N\REVISIONADD{eural}RTI-IT\\ (20)\end{tabular} &
  \cellcolor[HTML]{EFEFEF}PTM &
  \cellcolor[HTML]{EFEFEF}HSH 3 ord & 
  \cellcolor[HTML]{EFEFEF}PCA/RBF \\ 
  \hline
\cellcolor[HTML]{EFEFEF}Item 1  & 38.61/0.96 &38.13/0.96  & 38.47/0.96  & 36.92/0.92  & 37.01/0.92 & 35.01/0.98  & \textbf{39.66/0.98}  & 36.87/0.98 \\ \hline
\cellcolor[HTML]{EFEFEF}Item 2  & 36.49/0.95 & 36.53/0.95 & 36.42/0.95  & \textbf{38.89/0.97} & 38.50/0.97 & 27.66/0.96  & 37.88/0.98  & 32.11/0.96 \\ \hline
\cellcolor[HTML]{EFEFEF}Item 3  & 31.85/0.94 & 25.68/0.87 & 30.17/0.94  & \textbf{36.65/0.96} & 34.88/0.96 & 25.12/0.89  & 28.84/0.90   & 31.14/0.94 \\ \hline
\cellcolor[HTML]{EFEFEF}Item 4  & 33.49/0.95 & 30.17/0.92 & 30.86/0.91  & \textbf{37.38/0.97} & 36.81/0.97 & 25.29/0.95  & 32.18/0.98  & 32.70/0.98 \\ \hline
\cellcolor[HTML]{EFEFEF}Item 5  & 34.87/0.89 & 33.69/0.9 &  32.06/0.89 & \textbf{39.80/0.95} & 38.02/0.93 & 30.99/0.85  & 32.17/0.88  & 25.16/0.88 \\ \hline
\cellcolor[HTML]{EFEFEF}Item 6  & 38.64/0.95 & 38.63/0.95 & 37.50/0.94 & 38.72/0.95 & 34.27/0.92 & 33.64/0.93  & \textbf{39.69/0.96}  & 29.91/0.89 \\ \hline
\cellcolor[HTML]{EFEFEF}Item 7  & 29.97/0.92 & 16.8/0.68 & 26.82/0.92 & \textbf{39.17/0.95} & 38.84/0.95 & 32.32/0.97 & 30.06/0.96  & 29.28/0.95 \\ \hline
\cellcolor[HTML]{EFEFEF}Item 8  & 29.60/0.87 & 25.24/0.8 & 26.87/0.85  &  \textbf{35.96/0.92} & 35.49/0.91 & 29.43/0.89  & 29.75/0.90  & 27.92/0.88 \\ \hline
\cellcolor[HTML]{EFEFEF}Item 9  & 22.35/0.60 & 22.36/0.63 & 21.94/0.63  & 24.62/0.72 & \textbf{27.37/0.80} & 20.92/0.72  & 22.01/0.72  & 20.70/0.68 \\ \hline
\cellcolor[HTML]{EFEFEF}Item 10 & 23.21/0.74 & 22.45/0.71 & 23.00/0.75  & \textbf{30.54/0.90} & 29.25/0.89 & 16.90/0.60  & 19.92/0.66  & 17.99/0.65 \\ \hline
\cellcolor[HTML]{EFEFEF}Item 11 & 28.39/0.88 & 30.58/0.89 & 30.20/0.90  & 32.28/0.93 & \textbf{35.44/0.95} & 29.01/0.90  & 28.16/0.86  & 27.28/0.87 \\ \hline
\cellcolor[HTML]{EFEFEF}Item 12 & 31.33/0.90 & 30.53/0.89 &  31.31/0.89& \textbf{36.83/0.94} & 33.92/0.93 & 29.42/0.89  & 30.32/0.88 & 28.44/0.88 \\ \hline
 \cellcolor[HTML]{EFEFEF}Average &
\cellcolor[HTML]{EFEFEF}  31.56/0.88 &
\cellcolor[HTML]{EFEFEF}    29.23/0.85 &

\cellcolor[HTML]{EFEFEF}  30.46/0.88 & 
\cellcolor[HTML]{EFEFEF}  \textbf{35.65/0.92} &
\cellcolor[HTML]{EFEFEF}  34.98/0.93 &
\cellcolor[HTML]{EFEFEF}  27.95/0.88 &

\cellcolor[HTML]{EFEFEF}  30.77/0.88 &
\cellcolor[HTML]{EFEFEF}  28.29/0.87 \\ \hline
\end{tabular}%

\caption{Average PSNR/SSIM values for the relighting of test images of RealRTI collections. Values differ from \cite{Dulecha:2020:NRT} as we changed the testing protocol (see text). Figures in parentheses indicate the network layers' size.}
\label{tab:realRTI}
\end{table*}

\subsubsection{Student performance vs decoder size}
\label{sec:results:benchmarks:quality}
A first series of experiments was aimed at evaluating the effect of the Knowledge Distillation approach applied with the original network as the teacher and at finding a reasonable target size for the lightweight decoder.
\autoref{fig:line_synthrtiS}, \autoref{fig:line_synthrtiM}, and \autoref{fig:real_rti} represent the effect of the reduction of the decoder parameters on the quality of the relighting (measured by the PSNR of the comparison of relighted images and reference test images). Using the standard NeuralRTI training (blue lines), the quality becomes poor with the standard when the decoder layers are reduced from the original $50$ units ($3303$ decoder parameters) to 
\REVISIONADD{$40$ ($2243$),} $20$ ($723$) and $10$ ($263$) units.
For the DisK-N\REVISIONADD{eural}RTI training, instead, the results are still very good with $723$ parameters, even better than the original in the case of synthetic data  \REVISIONADD{(orange lines)}.
The decrease of the PSNR for smaller layers suggests that 20 can be a nearly optimal solution.
\autoref{tab:synthRTISingle}, \autoref{tab:synthRTIMulti}, and \autoref{tab:realRTI}
show the average PSNR and SSIM values of the comparisons between the ground truth test images and the ones relighted with different methods.
Looking at the first three columns, it is possible to appreciate the large improvement provided by the proposed training procedure based on knowledge distillation relative to the standard training of a decoder of the same size, also discussed in the conference paper~\cite{drp24}.

\begin{figure}[ht]
  \centering
  \begin{subfigure}[b]{0.32\linewidth}
    \includegraphics[width=\linewidth]{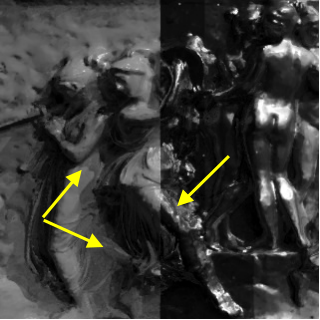}
    \caption{NeuralRTI (20)}
    \label{fig:comp_synth_mul_uncomp}
  \end{subfigure}  
    \begin{subfigure}[b]{0.32\linewidth}
    \includegraphics[width=\linewidth]{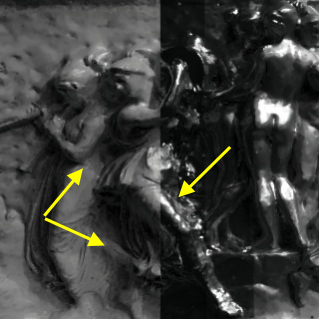}
    \caption{DisK-N\REVISIONADD{eural}RTI (20)}
    \label{fig:comp_synth_mul_comp}
  \end{subfigure}  
  \begin{subfigure}[b]{0.32\linewidth}
    \includegraphics[width=\linewidth]{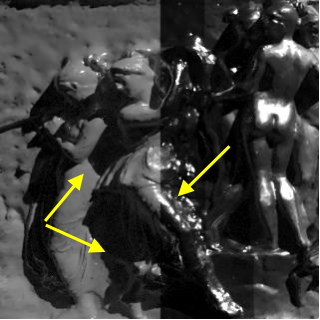}
    \caption{Ground Truth}
    \label{fig:comp_synth_mul_gt}
  \end{subfigure} 
  \caption{ (a) Relight with a test light direction of the SynthRTI multi-material set using the NeuralRTI(20) model. (b) Relight with the same light direction obtained with the DisK-NeuralRTI(20) compressed model. (c) Ground truth image corresponding to the test direction. It is possible to see (see arrows) that the layer size reduction with the original training (a) results in the loss of accuracy of the specular reflections and shadows. The image in (b) presents fewer artifacts compared with the ground truth (c). From \cite{drp24}.}
  \label{fig:comp_synth}
\end{figure}

%These results, see also our conference publication~\cite{drp24}, show the effectiveness of the Knowledge distillation approach.
\autoref{fig:comp_synth} shows an example of quality improvements deriving from the proposed compression strategy.

\subsubsection{Improvements of the relighting quality with the DisK\REVISIONADD{-Neural}RTI approach and the enhanced teacher}
Using the original ($3303$ parameters) and the lightweight ($723$) decoder, we performed extensive tests on the benchmarks, also increasing the complexity of the encoder/teacher as described in~\autoref{sec:method}.

The results presented in \autoref{tab:synthRTISingle}, \autoref{tab:synthRTIMulti}, and \autoref{tab:realRTI} show a comparison of the relighting quality obtained on the three low\REVISIONADD{-}resolution benchmarks\REVISIONADD{, using the quality metrics employed in the original NeuralRTI paper \cite{Dulecha:2020:NRT}.}

In particular, we compare the "classical" PTM, HSH and PCA/RBF methods with the original Neural RTI \cite{Dulecha:2020:NRT} trained with both original (NeuralRTI (50)) and smaller decoder (NeuralRTI (20)), the student network with reduced complexity trained with the original NeuralRTI architecture (DisK-N\REVISIONADD{eural}RTI), the Neural RTI version with additional layers (NeuralRTI-IT) and \REVISIONDEL{the }the student network with lighter decoder trained with this last version (DisK-N\REVISIONADD{eural}RTI-IT).

\begin{figure}[ht]
  \centering
  \begin{subfigure}[b]{0.48\linewidth}
    \includegraphics[width=\linewidth]{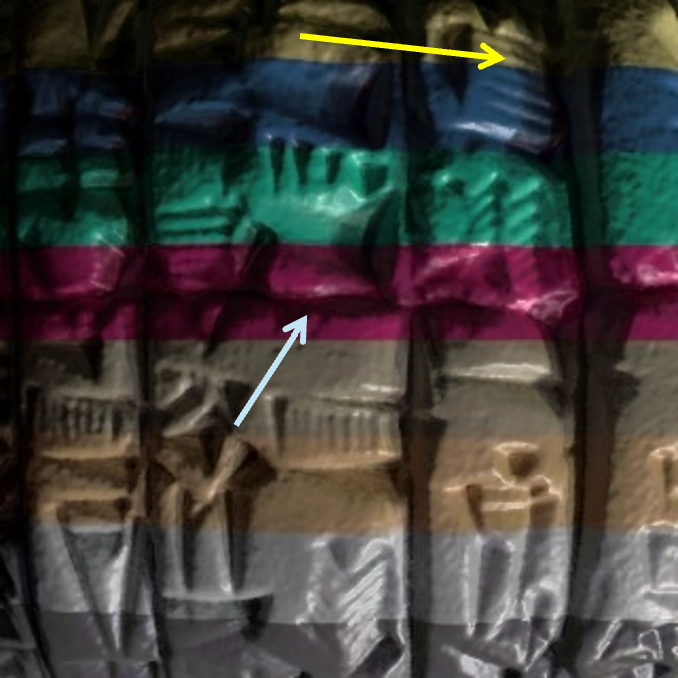}
    \caption{NeuralRTI (50)}
    \label{fig:mul_n50}
  \end{subfigure}  
    \begin{subfigure}[b]{0.48\linewidth}
    \includegraphics[width=\linewidth]{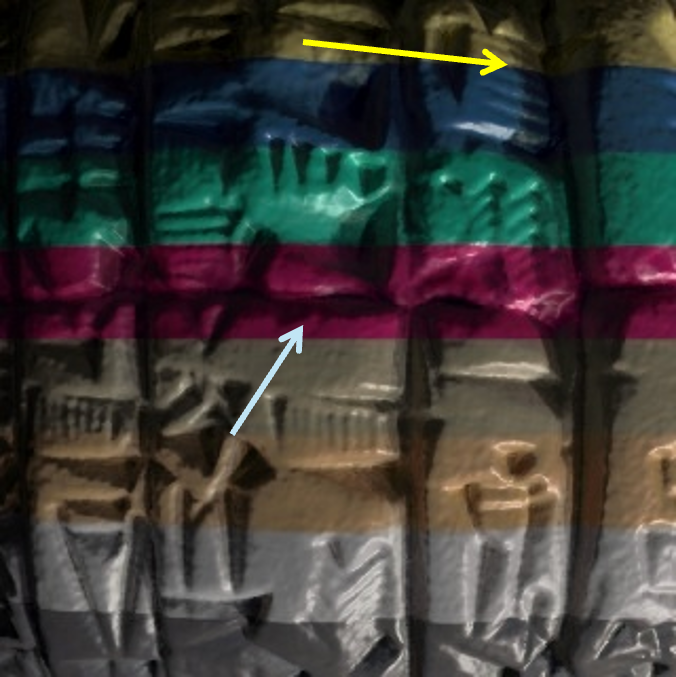}
    \caption{NeuralRTI (20)}
    \label{fig:mul_n20}
  \end{subfigure}  
  \begin{subfigure}[b]{0.48\linewidth}
    \includegraphics[width=\linewidth]{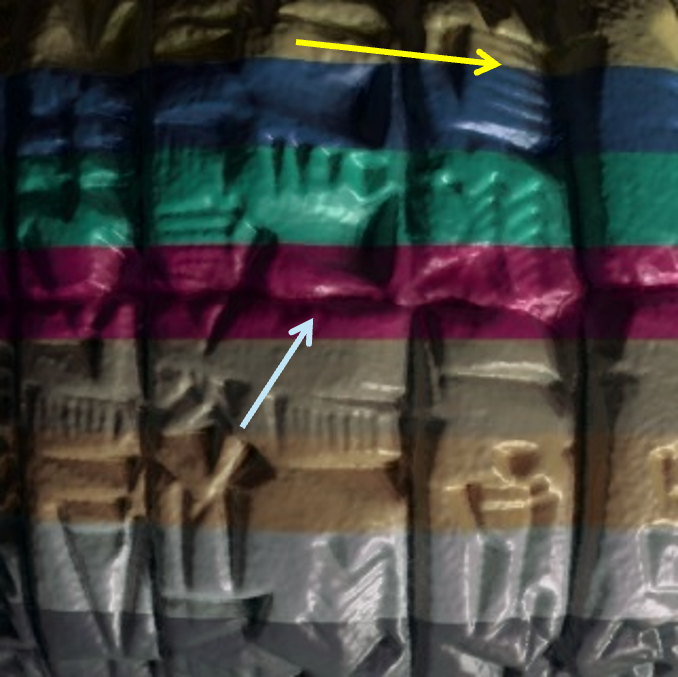}
    \caption{DisK-N\REVISIONADD{eural}RTI Improved (20)}
    \label{fig:mul_nim}
  \end{subfigure} 
    \begin{subfigure}[b]{0.48\linewidth}
    \includegraphics[width=\linewidth]{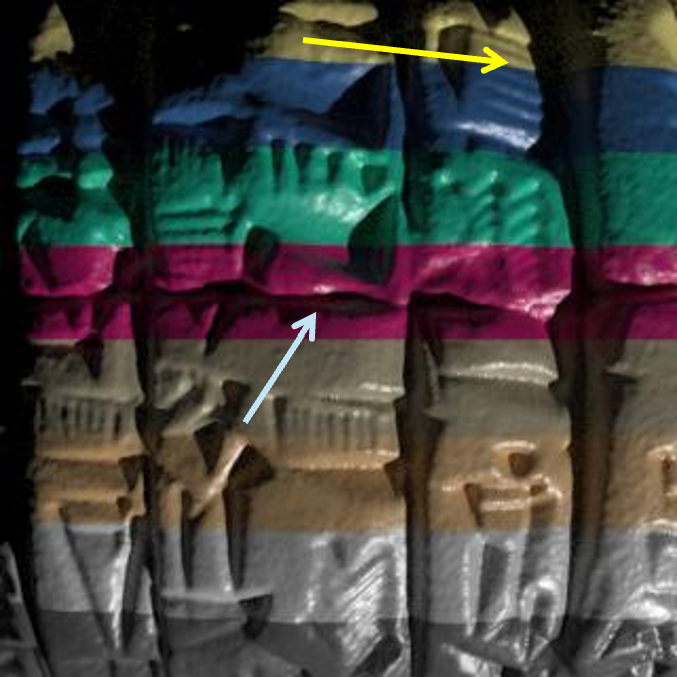}
    \caption{Ground Truth}
    \label{fig:mul_grt}
  \end{subfigure} 
  \caption{ (a) Relight with a test light direction of the SynthRTI multi-material set using the NeuralRTI (50) model. (b) Relight with the same light direction obtained with the NeuralRTI(20) compressed model with standard training. (c) Relight with the same light direction obtained with the NeuralRTI(20) compressed model trained with improved teacher and Knowledge Distillation. The last result is the only one avoiding artifacts in shadows (yellow arrow) and non-realistic highlight (cyan arrow) compared to the ground truth (d).}
  \label{fig:comp_synth2}
\end{figure}

The results consistently show that the increased encoder complexity gives strong advantages relative to the original one and that the training of the student network using knowledge distillation results in a very small decrease in the accuracy of the relight when the size of decoder layers is reduced to 20. The relighting with the DisK-N\REVISIONADD{eural}RTI-IT methods is much better than the original NeuralRTI, despite the decoder compression. Especially on the RealRTI benchmark, the reduction of the decoder size has a limited impact on the quality metrics.

The new results on RealRTI with the improved teacher also seem to demonstrate that, differently from what was achieved in our previous conference publication~\cite{drp24}, it is also possible to strongly enhance the accuracy of the relighting \REVISIONREP{relative}{with respect} to third-order HSH.

\autoref{fig:comp_synth2} shows a visual example of improvements obtained with the improved DisK-N\REVISIONADD{eural}RTI method relative to the original NeuralRTI with the heavier decoder. While the latter creates evident artifacts in the shadowed parts (highlighted by the yellow arrow) and cannot reproduce accurately small highlights (like the one indicated by the cyan arrow), the improved DisK-N\REVISIONADD{eural}RTI provides a result quite close to the ground truth without artifacts.

%DisK-NeuralRTI achieves comparable or superior results to the original method while using significantly fewer parameters. 
%The results are also better than all the classical methods, also those requiring a huge per-pixel encoding size.

\begin{figure*}[ht]
  \centering
  \begin{subfigure}[b]{0.24\linewidth}
    \includegraphics[width=\linewidth]{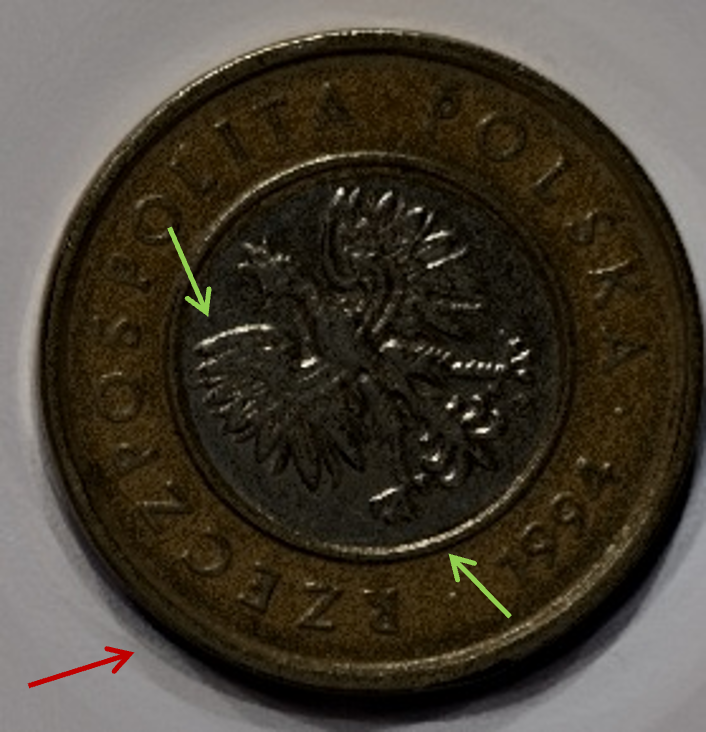}
    \caption{NeuralRTI}
    \label{fig:improved_real_mul_uncomp}
  \end{subfigure}  
    \begin{subfigure}[b]{0.24\linewidth}
    \includegraphics[width=\linewidth]{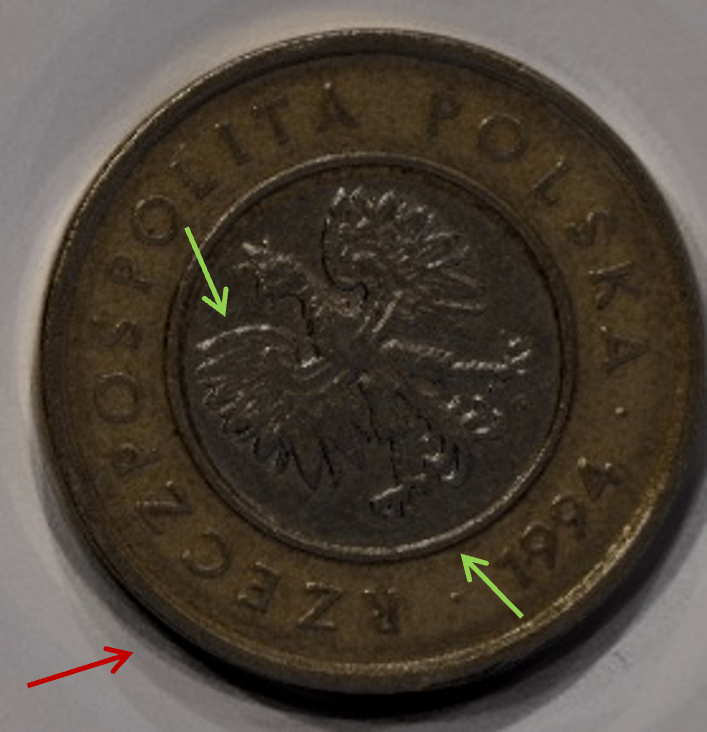}
    \caption{DisK-N\REVISIONADD{eural}RTI}
    \label{fig:improved_real_mul_comp}
  \end{subfigure}  
    \begin{subfigure}[b]{0.24\linewidth}
    \includegraphics[width=\linewidth]{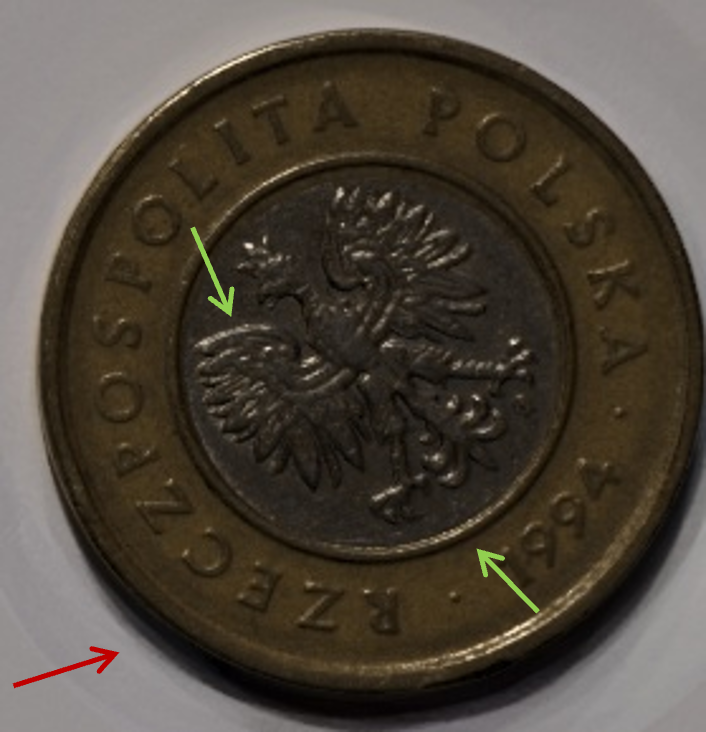}
    \caption{DisK-N\REVISIONADD{eural}RTI Improved}
    \label{fig:improved_real_mul_comp_imp}
  \end{subfigure}  
  \begin{subfigure}[b]{0.24\linewidth}
    \includegraphics[width=\linewidth]{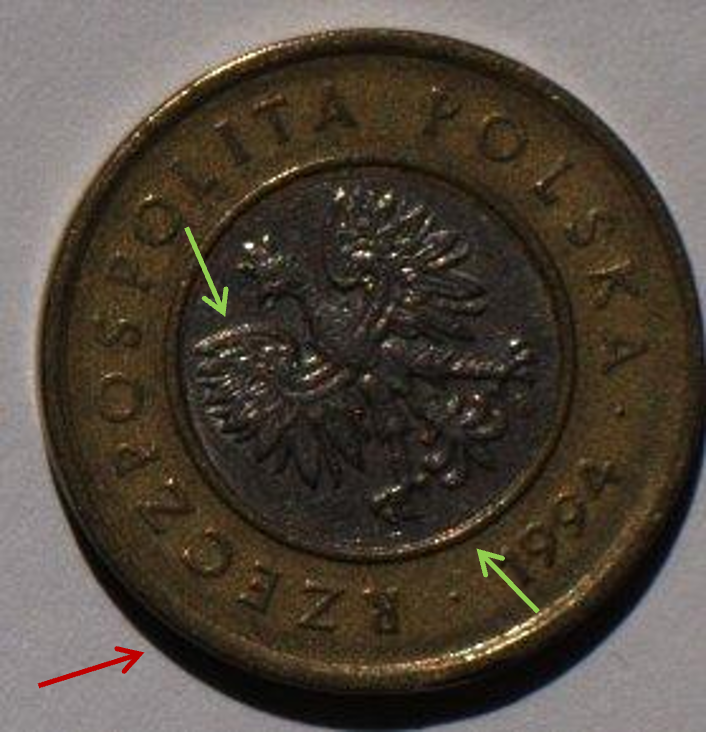}
    \caption{Ground Truth}
    \label{fig:improved_real_mul_gt}
  \end{subfigure} 
  \caption{Relight of a challenging object from the RealRTI benchmark. The relight obtained with the original Neural RTI method (a) reproduces the metallic behavior, but the golden part appears dark, the highlights are exaggerated with respect to the ground truth (d), and the cast shadow presents blending artifacts. Using this model to train a compressed decoder, we lose most of the highlights while the artifacts in the shadows are still there. The training of the lightweight decoder with the improved teacher, however, result in a relighted image with highlights and colors quite close to the ground truth and with reduced artifacts (c).}
  \label{fig:improved_real}
\end{figure*}

The better performances of the novel method, more evident on challenging objects with complex shapes and materials, are also illustrated in~\autoref{fig:improved_real}, showing relighted images from a test direction for item 9 of the RealRTI benchmark.
The original NeuralRTI with the heavy decoder and standard training fails in reproducing well the aspect of the golden metal and presents evident blending artifacts in the coin's shadow. Applying the DisK-NeuralRTI method, keeping the original network as the teacher, seems to improve the aspect of the golden metallic areas, but provides a poorer representation of the highlights. The architecture with a lightweight decoder trained with the improved teacher results in a better reproduction of material properties and highlights and removes the evident artifacts in the shadow (\autoref{fig:improved_real}).
Using the DisK-NeuralRTI with the improved teacher, it is possible to strongly improve the results obtained with the original model and reproduce quite well the complex material behavior of metallic surfaces (see the accurate reproduction of the highlights indicated by the green arrows). An evident improvement is also seen in the absence of blending artifacts in the cast shadow (indicated by the red arrows).

\REVISIONADD{ The better perceived quality of the NeuralRTI-based relighting with the novel teacher, and the effective compression obtained with the Knowledge Distillation-based approach are also evident looking at the comparisons performed using LPIPS and DeltaE, reported in~\mbox{\autoref{tab:synthRTISingleNM}}, \mbox{\autoref{tab:synthRTIMultiNM}}, and \mbox{\autoref{tab:realRTINM}}.}

\REVISIONADD{
In particular, it is possible to see that the network with the improved teacher provides, on average, a large improvement in the metrics with respect to the classical methods and the original NeuralRTI. 
The compression with DisK-NeuralRTI results in a non-negligible decrease of the LPIPS, which remains, however, comparable with the original network on SynthRTI and significantly better than the original NeuralRTI on RealRTI. 
In the case of DeltaE, measuring the preservation of the chromaticity in the relighting, the Neural method works quite well even though it does not process separately the color channels like PTM and HSH, with the improved teacher providing much better scores and, surprisingly, negligible worsening, or even improvements (on RealRTI) after the compression.
}

\begin{table*}[ht]
\footnotesize
\centering
\setlength{\tabcolsep}{2pt}

\begin{REVISIONADDENV}
\begin{tabular}{|l|c|c|c|c|c|c|c|c|}
\hline
  \cellcolor[HTML]{EFEFEF}&
  \cellcolor[HTML]{EFEFEF}\begin{tabular}[c]{@{}c@{}}NeuralRTI\\ (50)\end{tabular} & 
  \cellcolor[HTML]{EFEFEF}\begin{tabular}[c]{@{}c@{}}NeuralRTI\\ (20)\end{tabular} & 
    \cellcolor[HTML]{EFEFEF}\begin{tabular}[c]{@{}c@{}}DisK-NeuralRTI\\ (20)\end{tabular} &
  \cellcolor[HTML]{EFEFEF}\begin{tabular}[c]{@{}c@{}}NeuralRTI-IT\\ (50)\end{tabular} &

  \cellcolor[HTML]{EFEFEF}\begin{tabular}[c]{@{}c@{}}DisK-NeuralRTI-IT\\ (20)\end{tabular} &
  \cellcolor[HTML]{EFEFEF}PTM &
  
  \cellcolor[HTML]{EFEFEF}HSH 3ord &
  \cellcolor[HTML]{EFEFEF}PCA/RBF \\ \hline
\cellcolor[HTML]{EFEFEF}Canvas & 0.019/0.88  & 0.027/1.06 & 0.020/0.89 & \textbf{0.014/0.39} & 0.019/0.48  & 0.075/3.23 & 0.036/0.91 & 0.038/2.63 \\ \hline
\cellcolor[HTML]{EFEFEF}Tablet & 0.098/2.45 & 0.110/2.71   & 0.124/2.72 & \textbf{0.065}/2.32 & 0.094/\textbf{2.23} & 0.188/4.15 & 0.111/2.48 & 0.167/3.73 \\ \hline
\cellcolor[HTML]{EFEFEF}Bas-relief  & 0.080/2.4  & 0.091/2.37 & 0.103/2.39 & \textbf{0.039/1.40}  & 0.080/1.95  & 0.171/4.53 & 0.091/2.32 & 0.153/3.77 \\ \hline
\rowcolor[HTML]{EFEFEF} 
Average                         & 0.065/1.91 & 0.076/2.05 & 0.082/2.00 & \textbf{0.039/1.37} & 0.064/1.55 & 0.145/3.97 & 0.079/1.90 & 0.119/3.38 \\ \hline

\end{tabular}
\end{REVISIONADDENV}
\caption{\label{tab:synthRTISingleNM} \REVISIONADD{Average LPIPS/DeltaE values for the relighting of test images of SynthRTI SingleMaterial collections.  Bold figures indicate the best values. Figures in parentheses indicate the network layers’ size.}}
\end{table*}

\begin{table*}[ht]
\footnotesize
\centering
\setlength{\tabcolsep}{2pt}
\begin{REVISIONADDENV}
\begin{tabular}{|l|c|c|c|c|c|c|c|c|}
\hline
 \cellcolor[HTML]{EFEFEF}&
  \cellcolor[HTML]{EFEFEF}\begin{tabular}[c]{@{}c@{}}NeuralRTI\\ (50)\end{tabular} &
  \cellcolor[HTML]{EFEFEF}\begin{tabular}[c]{@{}c@{}}NeuralRTI\\  (20)\end{tabular} &
  \cellcolor[HTML]{EFEFEF}\begin{tabular}[c]{@{}c@{}}DisK-NeuralRTI\\ (20)\end{tabular} &
    \cellcolor[HTML]{EFEFEF}\begin{tabular}[c]{@{}c@{}}NeuralRTI-IT\\  (50)\end{tabular} &
  \cellcolor[HTML]{EFEFEF}\begin{tabular}[c]{@{}c@{}}DisK-NeuralRTI-IT\\ (20)\end{tabular} &
  \cellcolor[HTML]{EFEFEF}PTM &
  \cellcolor[HTML]{EFEFEF}HSH 3 ord &
  \cellcolor[HTML]{EFEFEF}PCA/RBF \\ \hline
\cellcolor[HTML]{EFEFEF}Canvas & 0.036/2.30  & 0.046/2.69    & 0.041/2.48         & \textbf{0.022/1.65} & 0.038/2.35          & 0.092/5.08 & 0.065/2.80  & 0.068/4.03 \\ \hline
\cellcolor[HTML]{EFEFEF}Tablet & 0.120/4.89 & 0.185/6.73 & 0.147/5.24 & \textbf{0.101}/5.23 & 0.119/4.56 & 0.217/6.53 & 0.129/\textbf{4.26} & 0.237/7.11 \\ \hline
\cellcolor[HTML]{EFEFEF}Bas-relief  & 0.097/3.95 & 0.115/4.34    & 0.120/4.28          & \textbf{0.076/3.71} & 0.106/3.78          & 0.202/5.62 & 0.125/3.87 & 0.23/6.13  \\ \hline
\rowcolor[HTML]{EFEFEF} 
Average                         & 0.084/3.71 & 0.115/4.59    & 0.103/4.00           & \textbf{0.066/3.53} & 0.088/3.56          & 0.17/5.74  & 0.106/3.64 & 0.178/5.76 \\ \hline

\end{tabular}
\end{REVISIONADDENV}
\caption{\label{tab:synthRTIMultiNM} \REVISIONADD{Average LPIPS/DeltaE values for the relighting of test images of SynthRTI MultiMaterial collections. Bold figures indicate the best values. Figures in parentheses indicate the network layers’ size.}}
\end{table*}

\begin{table*}[ht]

\footnotesize
\centering
\setlength{\tabcolsep}{2pt}
\begin{REVISIONADDENV}
\begin{tabular}{|l|c|c|c|c|c|c|c|c|}
\hline
 \cellcolor[HTML]{EFEFEF}&
  \cellcolor[HTML]{EFEFEF}\begin{tabular}[c]{@{}c@{}}NeuralRTI\\ (50)\end{tabular} &
    \cellcolor[HTML]{EFEFEF}\begin{tabular}[c]{@{}c@{}}NeuralRTI\\ (20)\end{tabular} &
      \cellcolor[HTML]{EFEFEF}\begin{tabular}[c]{@{}c@{}}DisK-NeuralRTI\\ (20)\end{tabular} &
  \cellcolor[HTML]{EFEFEF}\begin{tabular}[c]{@{}c@{}}NeuralRTI-IT\\  (50)\end{tabular} &
  \cellcolor[HTML]{EFEFEF}\begin{tabular}[c]{@{}c@{}}DisK-NeuralRTI-IT\\ (20)\end{tabular} &
  \cellcolor[HTML]{EFEFEF}PTM &
  \cellcolor[HTML]{EFEFEF}HSH 3 ord & 
  \cellcolor[HTML]{EFEFEF}PCA/RBF \\ 
  \hline
\cellcolor[HTML]{EFEFEF}Item 1 & 0.017/1.59 & 0.027/1.85  & 0.022/1.60 & \textbf{0.017/1.88} & 0.019/1.94 & 0.078/6.60  & 0.079/6.76  & 0.020/1.90 \\ \hline
\cellcolor[HTML]{EFEFEF}Item 2 & 0.032/1.70 & 0.043/1.77  & 0.039/1.79 & 0.029/1.42  & 0.037/1.57 & 0.045/3.49  & \textbf{0.021/1.41}  & 0.054/2.86 \\ \hline
\cellcolor[HTML]{EFEFEF}Item 3 & 0.077/4.92 & 0.080/4.56  & 0.064/3.65 & \textbf{0.024/2.08} & 0.030/2.37 & 0.160/12.71   & 0.161/13.93 & 0.058/3.25  \\ \hline
\cellcolor[HTML]{EFEFEF}Item 4 & 0.025/3.03 & 0.033/3.52  & 0.028/3.60 & \textbf{0.012/1.81}  & 0.015/2.04 & 0.110/16.73 & 0.086/12.60 & 0.019/2.48 \\ \hline
\cellcolor[HTML]{EFEFEF}Item 5 & 0.062/2.06 & 0.075/2.38  & 0.077/2.57 & \textbf{0.042/1.52} & 0.058/1.76  & 0.091/2.57  & 0.104/2.49  & 0.092/4.94 \\ \hline
\cellcolor[HTML]{EFEFEF}Item 6 & 0.041/1.60 & 0.040/1.46  & 0.048/1.58 & 0.035/1.44 & \textbf{0.035}/1.47 & 0.055/2.06  & \textbf{0.027/1.35}  & 0.069/3.13 \\ \hline
\cellcolor[HTML]{EFEFEF}Item 7 & 0.072/3.89 & 0.712/15.00 & 0.083/5.79 & \textbf{0.037/1.07} & 0.048/1.24 & 0.070/2.67  & 0.045/2.60  & 0.095/3.75 \\ \hline
\cellcolor[HTML]{EFEFEF}Item 8 & 0.102/4.39  & 0.122/7.03   & 0.129/5.87 & 0.067/\textbf{1.76} & 0.076/1.83  & 0.095/3.21  & \textbf{0.062}/2.54  & 0.128/3.75 \\ \hline
\cellcolor[HTML]{EFEFEF}Item 9 & 0.114/5.99 & 0.137/6.04  & 0.138/6.43 & \textbf{0.098}/6.12 & 0.103/\textbf{4.02} & 0.210/6.83  & 0.138/6.05  & 0.180/7.03 \\ \hline
\cellcolor[HTML]{EFEFEF}Item 10 & 0.147/4.96 & 0.163/5.48  & 0.142/5.12 & \textbf{0.078/3.46} & 0.089/3.72 & 0.268/10.80  & 0.170/7.05  & 0.217/8.95 \\ \hline
\cellcolor[HTML]{EFEFEF}Item 11 & 0.093/3.59 & 0.089/2.69  & 0.084/2.61 & 0.055/2.50 & \textbf{0.049/2.06} & 0.184/7.15  & 0.174/7.02  & 0.138/3.66 \\ \hline
\cellcolor[HTML]{EFEFEF}Item 12 & 0.158/2.35 & 0.188/2.52  & 0.196/2.42 & \textbf{0.086/1.71} & 0.155/2.06  & 0.214/2.60  & 0.168/2.37  & 0.258/3.11 \\ \hline
\rowcolor[HTML]{EFEFEF} 
Average                        & 0.078/3.34 & 0.142/4.53  & 0.088/3.59 & \textbf{0.048}/2.23 & 0.060/\textbf{2.17} & 0.132/6.45  & 0.103/5.51   & 0.111/4.07 \\ \hline
\end{tabular}
\end{REVISIONADDENV}
\caption{\label{tab:realRTINM} \REVISIONADD{Average LPIPS/DeltaE values for the relighting of test images of RealRTI collections. 
Bold figures indicate the best values. Figures in parentheses indicate the network layers’ size.}}
\end{table*}

\begin{figure*}[ht]
  \centering
    \begin{subfigure}[b]{0.445\linewidth}
    \includegraphics[height=4.2cm]{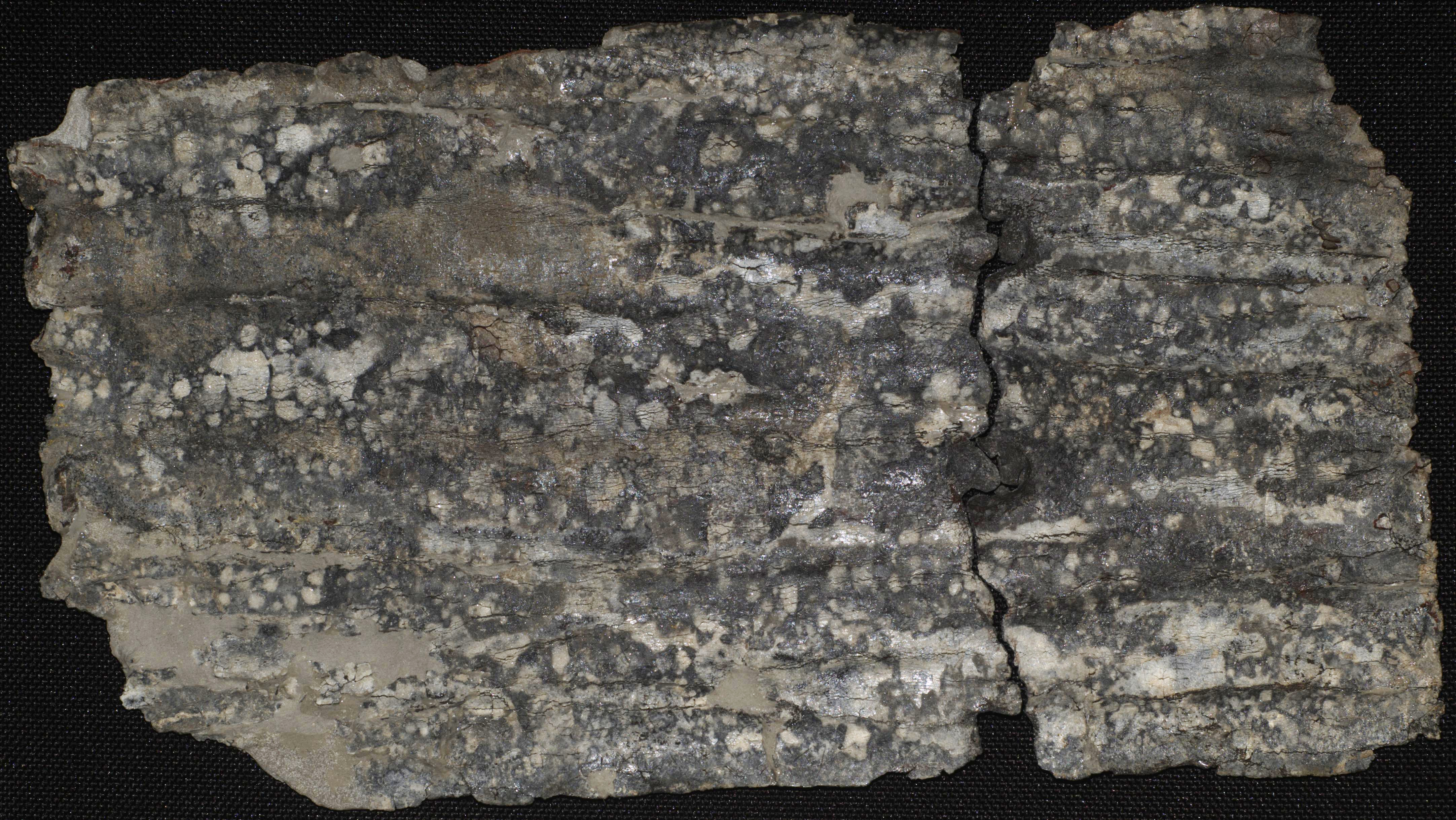}
    \caption{Lamina}
    \label{fig:lamina}
  \end{subfigure}  
  \begin{subfigure}[b]{0.33\linewidth}
    \includegraphics[height=4.2cm]{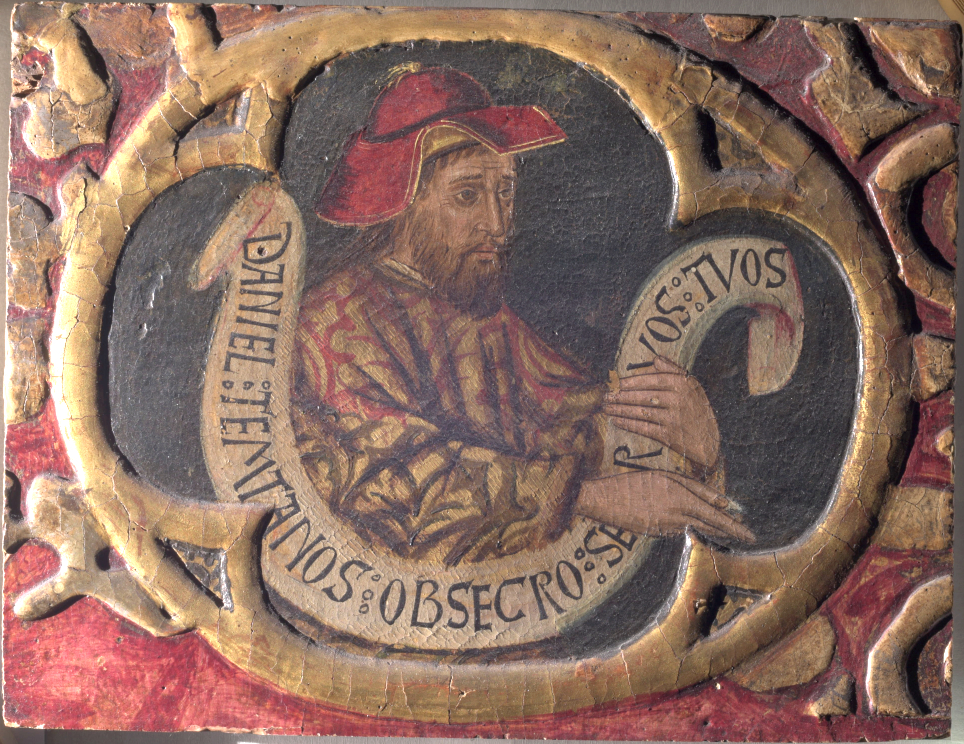}
    \caption{Retablo (small)}
    \label{fig:retable_small}
  \end{subfigure}  
  \begin{subfigure}[b]{0.21\linewidth}
    \includegraphics[height=4.2cm]{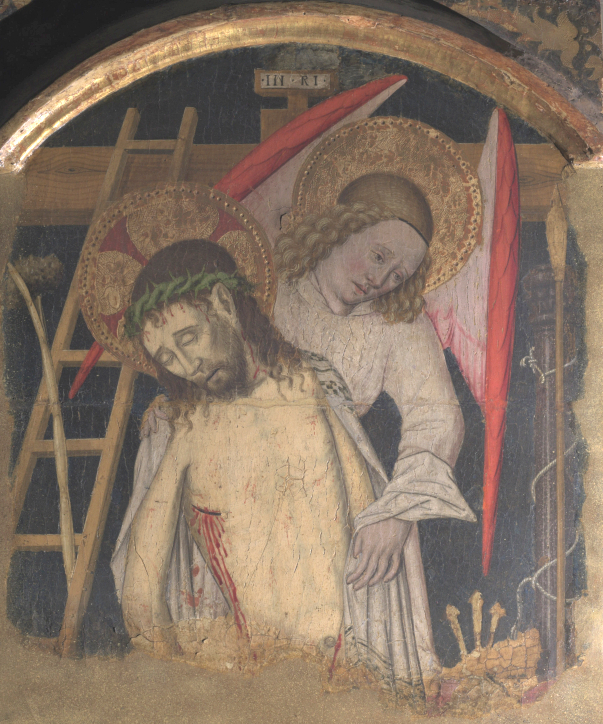}
    \caption{Retablo (big)}
    \label{fig:retable_big}
  \end{subfigure}

  \begin{subfigure}[b]{0.32\linewidth}    \REVISIONIMG{\includegraphics[height=3.9cm]{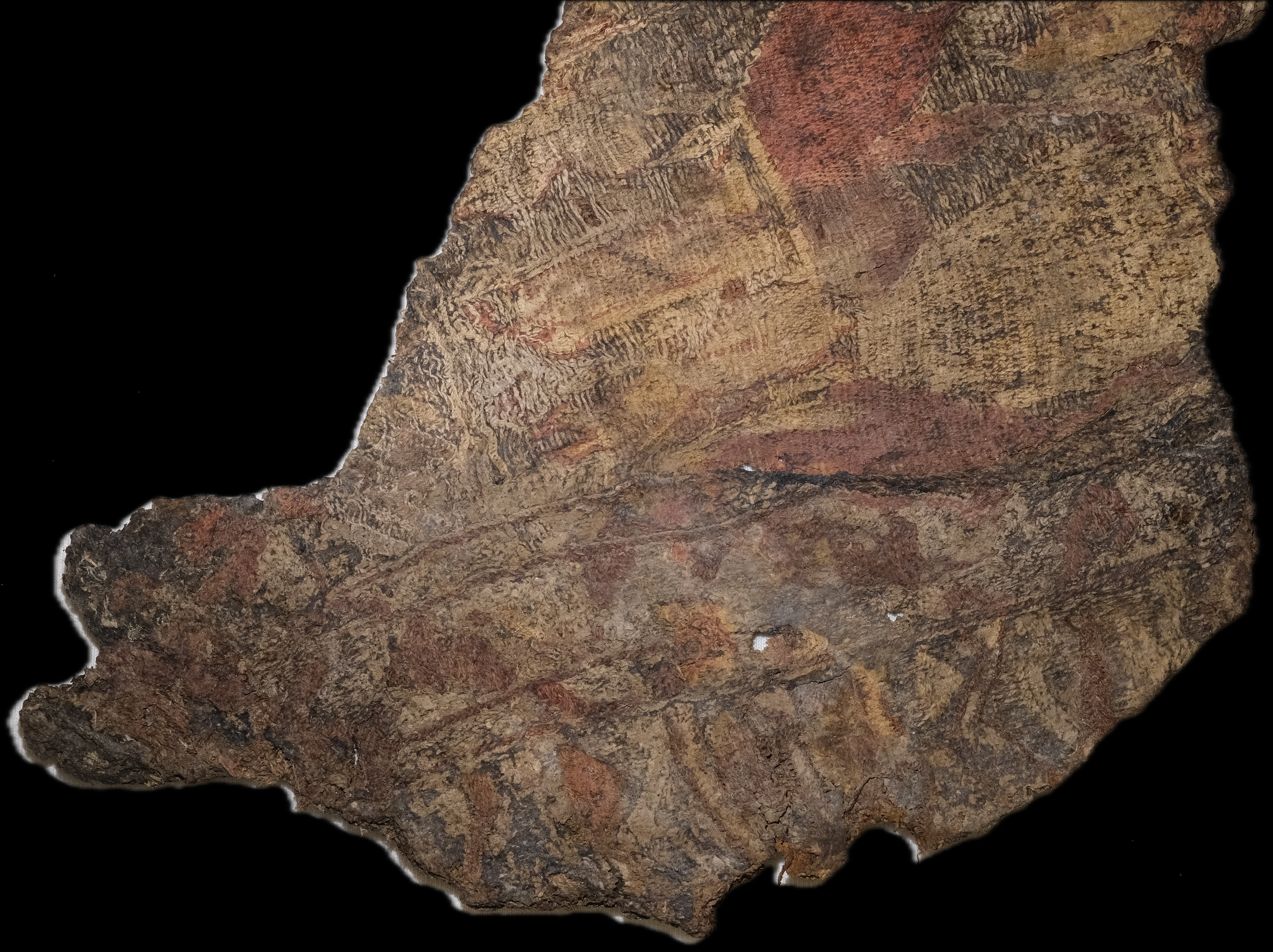}}
    \caption{\REVISIONADD{Textile fragment}}
    \label{fig:textile_fragment}
  \end{subfigure}  
  \begin{subfigure}[b]{0.67\linewidth}
    \REVISIONIMG{\includegraphics[height=3.9cm]{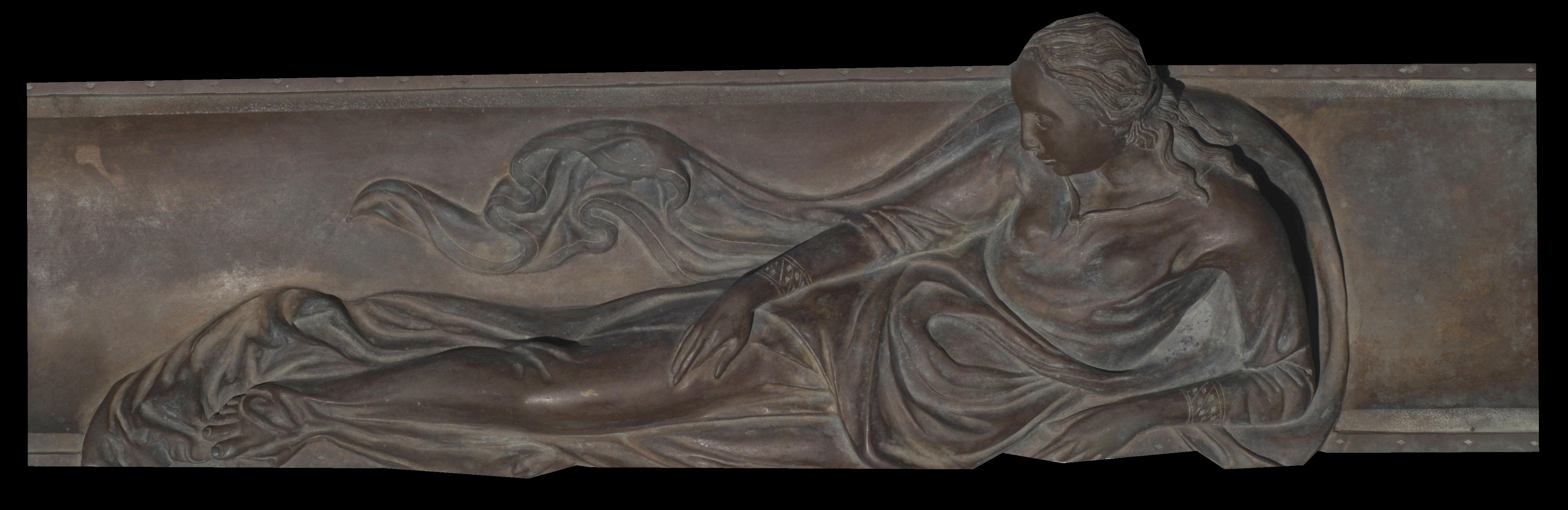}}
    \caption{\REVISIONADD{Relieved bronze panel}}
    \label{fig:bronze_panel}
  \end{subfigure} 
  
  \caption{Example images of the real-world Cultural Heritage MLICs used for benchmarking. (a): lead sheet found in Cesarea Marittima, Israel. (b), (c): Panels from the retable of St. Bernardino (1455), Cagliari, Italy. \REVISIONADD{(d); Textile fragment from the Oseberg find. (e) Relieved bronze panel, copy of Lorenzo Ghiberti’s Paradise Door.}}
  \label{fig:ch-datasets}
\end{figure*}

\section{Evaluation of high resolution image relighting and the RealRTIHR dataset.}
\label{sec:results:chmlic}
For the evaluation of the relighting in a realistic application setting, we created a novel dataset with MLICs made of large images coming from real studies of cultural heritage artifacts featuring different material properties and shape complexity.
On these images, it is possible not only to evaluate the objective quality of the relighted images, but also to test the interactive performance of the viewer application and the potential issues related to the training of the network.

\subsection{The RealRTIHR dataset}
The dataset is composed of high-resolution MLICs captured in \REVISIONREP{across multiple}{in different} cultural heritage projects for \REVISIONREP{different purposes}{the analysis of surface properties}, featuring different \REVISIONADD{shape and } material \REVISIONREP{characteristics}{properties}:
\begin{itemize}
    \item A \textbf{lead sheet} found in the 1960s in Caesarea Maritima, during the excavations of an Italian archaeological mission\REVISIONADD{~(\mbox{\autoref{fig:lamina}})}.
    The item is now at the Archaeological Museum of Milan and was acquired to study the engraved inscriptions. The metallic surface is not flat and presents different degrees of roughness. The MLIC data were obtained with a light dome (47 LED) and a Nikon D810 DSLR camera. Original images were cropped to a resolution equal to 4328 × 2436 (10.5Mp).

    \item \textbf{A small panel} ($34 \times 25$ cm.) from the retable of St. Bernardino, painted in oil on a wooden support and dated 1455\REVISIONADD{~(\mbox{\autoref{fig:retable_small}})}.
   The item is now housed and displayed at the Pinacoteca Nazionale in Cagliari. The surface features different brilliant colors, variations in shininess, and relieved structures.
   The MLIC data have been captured with a 36.3 Megapixel DSLR FX Nikon D810 Camera with a 50 AF Nikkor Lens and a handheld white LED (5500K) that spans the entire visible spectrum. Images were cropped to 3811x2451 (9.34Mp).
   
   \item \textbf{A larger panel}  ($54 \times 36$ cm.) from the same polyptych, featuring a golden arched frame with an image of Christ in pity, supported by an angel~. It has been acquired with the same setup as the previous one.  Images were cropped to 4117x3427 (14.1Mp).

\begin{REVISIONADDENV}
   \item \textbf{An ancient textile fragment}  coming from a Viking Age burial mound at Oseberg in south Norway~(\mbox{\autoref{fig:textile_fragment}}). Data is courtesy of Tomasz Łojewski (AGH University of Science and Technology, Kraków). The interesting aspect of the surface is the presence of fine patterns creating shadows in the matte surface of the tissue. Images have a resolution of 6240x4160 (25.9 Mp).

   \item \textbf{A bronze panel} representing a female figure. It is a copy of a bronze panel of Lorenzo Ghiberti’s Paradise Door of the Florence Baptistery~(\mbox{\autoref{fig:bronze_panel}}). The item was cast with a Cu90-Sn10 alloy, a type of bronze with very good corrosion resistance and durability. An artificial patination was applied to the surface by using Iron (III) Chloride to give the surface a brownish appearance. The item has been created for the Scan4Reco European project~\cite{giachetti2017multispectral,Giachetti:2018:NFH}
   The MLIC data have been captured with a 36.3 Megapixel Nikon D810 DSLR FX Nikon D810 Camera and a Handheld light and have been cropped to a resolution 4576x1488 (6.8Mp).
\end{REVISIONADDENV}
\end{itemize}

For the benchmarking of novel view generation, we split the original MLIC data into a training and test set. %selecting test light directions with different elevation.
%to test the quality of the relighting on the new high-resolution images.  
%
This was done keeping an approximately uniform sampling in the train data and separating a minimum of 5 images with varying elevation for the test set.
%the light directions with the smaller angular distance from the directions of the virtual dome of the SynthRTI Train dataset. The test set is generated by selecting images with variable elevation.

\begin{table*}[ht]
\footnotesize
\setlength{\tabcolsep}{2pt}
\centering
\begin{tabular}{|l|c|c|c|c|c|c|c|c|c|}
\hline
 \cellcolor[HTML]{EFEFEF}&
  \cellcolor[HTML]{EFEFEF}\begin{tabular}[c]{@{}c@{}}NeuralRTI\\ (50)\end{tabular} &
  \cellcolor[HTML]{EFEFEF}\begin{tabular}[c]{@{}c@{}}NeuralRTI\\  (20)\end{tabular} &
  \cellcolor[HTML]{EFEFEF}\begin{tabular}[c]{@{}c@{}}DisK-N\REVISIONADD{eural}RTI\\ (20)\end{tabular} &
     \cellcolor[HTML]{EFEFEF}\begin{tabular}[c]{@{}c@{}}NeuralRTI-IT\\  (50)\end{tabular} &
  \cellcolor[HTML]{EFEFEF}\begin{tabular}[c]{@{}c@{}}DisK-N\REVISIONADD{eural}RTI-IT\\ (20)\end{tabular} &
  
  \cellcolor[HTML]{EFEFEF}PTM &
  \cellcolor[HTML]{EFEFEF}HSH 3 ord &
  \cellcolor[HTML]{EFEFEF}PCA/RBF \\ \hline
\cellcolor[HTML]{EFEFEF}Lamina         & 37.29/0.92 & 33.00/0.83    & 36.47/0.94    & 37.57/0.92  &   \textbf{38.33/0.94}   & 32.35/0.86   & 34.53/0.88 & 31.50/0.84  \\ \hline
\cellcolor[HTML]{EFEFEF}Retablo\_small & 31.68/0.84 & 26.84/0.72 & 31.04/0.82      &  \textbf{32.10/0.85} &   31.07/0.82 & 23.06/0.76 &  24.92/0.77 & 24.34/0.76 \\ \hline
\cellcolor[HTML]{EFEFEF}Retablo\_big   & 37.57/0.95          & 33.37/0.92 & 38.33/0.95 & \textbf{38.40/0.96} & 36.44/0.95&27.99/0.92  & 29.54/0.92 & 29.14/0.92 \\ \hline
\cellcolor[HTML]{EFEFEF}Textile fragment           & 29.94/0.92 & 29.49/0.90 & 30.08/0.92 & \textbf{31.25/0.94} & 31.11/\textbf{0.94} & 29.95/0.92 & 30.36/0.92 & 29.61/0.91 \\ \hline
\cellcolor[HTML]{EFEFEF}Bronze panel      & 35.57/0.93 & 33.95/0.92 & 34.37/0.92 & \textbf{35.52}/0.93 & 33.80/\textbf{0.94} & 32.96/0.92 & 32.54/0.89 & 32.18/0.91 \\ \hline
\rowcolor[HTML]{EFEFEF} 
Average &
  \cellcolor[HTML]{EFEFEF}34.41/0.91 &
  \cellcolor[HTML]{EFEFEF}31.33/0.86 &
  34.06/0.91 &
  \cellcolor[HTML]{EFEFEF}\textbf{35.06/0.92} &
  34.24/\textbf{0.92} &
  \cellcolor[HTML]{EFEFEF}29.22/0.85 &
  30.34/0.85 &
  29.35/0.85 \\ \hline
\rowcolor[HTML]{EFEFEF} 
\end{tabular}%
\caption{\label{tab:realrti2}Average PSNR/SSIM of the methods on the different high-resolution datasets. The quality of the neural relight is far better with the neural model \REVISIONREP{relative}{with respect} to classical techniques. and the compression with DisK-N\REVISIONADD{eural}RTI \REVISIONREP{improves performance to a level supporting interactivity while not affecting the rendering quality}{does not affect the quality}.}
\end{table*}

\begin{table*}[ht]
\footnotesize
\setlength{\tabcolsep}{2pt}
\centering
\begin{REVISIONADDENV}
\begin{tabular}{|l|c|c|c|c|c|c|c|c|c|}
\hline
 \cellcolor[HTML]{EFEFEF}&
  \cellcolor[HTML]{EFEFEF}\begin{tabular}[c]{@{}c@{}}NeuralRTI\\ (50)\end{tabular} &
  \cellcolor[HTML]{EFEFEF}\begin{tabular}[c]{@{}c@{}}NeuralRTI\\  (20)\end{tabular} &
  \cellcolor[HTML]{EFEFEF}\begin{tabular}[c]{@{}c@{}}DisK-NeuralRTI\\ (20)\end{tabular} &
     \cellcolor[HTML]{EFEFEF}\begin{tabular}[c]{@{}c@{}}NeuralRTI-IT\\  (50)\end{tabular} &
  \cellcolor[HTML]{EFEFEF}\begin{tabular}[c]{@{}c@{}}DisK-NeuralRTI-IT\\ (20)\end{tabular} &
  
  \cellcolor[HTML]{EFEFEF}PTM &
  \cellcolor[HTML]{EFEFEF}HSH 3 ord &
  \cellcolor[HTML]{EFEFEF}PCA/RBF \\ \hline
\cellcolor[HTML]{EFEFEF}Lamina         & \textbf{0.014/1.11} & 0.049/1.46 & 0.016/1.17 & 0.018/1.25 & 0.022/1.12 & 0.047/1.68 & 0.024/1.34 & 0.099/1.84 \\ \hline
\cellcolor[HTML]{EFEFEF}Retablo\_small & 0.014/3.61 & 0.030/4.84 & 0.017/3.80 & \textbf{0.012/3.50} & 0.017/3.81 & 0.113/6.73 & 0.065/5.71 & 0.103/6.74 \\ \hline
\cellcolor[HTML]{EFEFEF}Retablo\_big   & 0.010/2.01 & 0.021/2.34 & 0.014/1.93 & \textbf{0.006/1.61} & 0.017/2.00 & 0.059/3.45 & 0.038/2.89 & 0.066/3.48 \\ \hline
\cellcolor[HTML]{EFEFEF}Textile fragment          & 0.021/4.17 & 0.031/5.07 & 0.023/4.20 & \textbf{0.015/3.44} & 0.020/3.54 & \textbf{0.015}/3.95 & 0.006/3.50 & 0.030/4.88 \\ \hline
\cellcolor[HTML]{EFEFEF}Bronze panel      & 0.031/2.72 & 0.035/2.77 & 0.043/2.79 & 0.030/\textbf{2.58} & \textbf{0.029}/2.82 & 0.059/2.79 & 0.048/3.17 & 0.072/3.64 \\ \hline
\rowcolor[HTML]{EFEFEF} 
\cellcolor[HTML]{EFEFEF}Average &
  0.018/2.72 &
  0.033/3.30 &
  0.023/2.78 &
  \textbf{0.016/2.48} &
  0.021/2.66 &
  0.059/3.72 &
  0.036/3.32 &
  \cellcolor[HTML]{EFEFEF}0.074/4.12 \\ \hline
\end{tabular}%
\end{REVISIONADDENV}
\caption{\label{tab:realrti2NM}\REVISIONADD{Average LPIPS/DeltaE of the methods on the different high-resolution datasets.}}
\end{table*}

\subsubsection{Relighting quality evaluation on RealRTIHR}
\label{sec:results:chmlic:quality}

\REVISIONREP{We used the training sets to fit all the classical and neural relighting models, and the test light directions to create the novel images 
to be compared against the ground-truth images, exactly as done for the SynthRTI and RealRTI benchmarks.}{We used the training sets to fit all the classical and neural relighting image models and the test light directions to create the novel images and evaluate the PSNR and SSIM values of their comparisons against the ground-truth images, exactly as done for the SynthRTI and RealRTI benchmarks.}

PSNR and SSIM scores obtained with the different techniques are reported in 
\autoref{tab:realrti2}\REVISIONADD{, while LPIPS and DeltaE values are shown in \autoref{tab:realrti2NM}.}

The quality provided by the neural methods is consistently better than that provided by the classical ones. 

The average PSNR obtained with our method is approximately 20\% higher than the one obtained with third-order HSH which is a huge difference. \REVISIONADD{The average perceptual loss (LPIPS) provided by Neural-RTI-IT is halved compared to the corresponding result obtained with third-order HSH.}

The improved teacher (IT) allows the method to \REVISIONREP{enhance}{improve} the results obtained by the previous architecture, and the metrics obtained with the lightweight decoder with layers made of 20 elements, trained with the knowledge distillation approach from the improved teacher (DisK-N\REVISIONADD{eural}RTI-IT) are \REVISIONREP{close}{superior} to those obtained with the original NeuralRTI\REVISIONADD{, and even better for SSIM and DeltaE}. 

\begin{figure*}[ht]
  \centering
  \begin{subfigure}[b]{0.32\linewidth}
  \includegraphics[width=\linewidth]{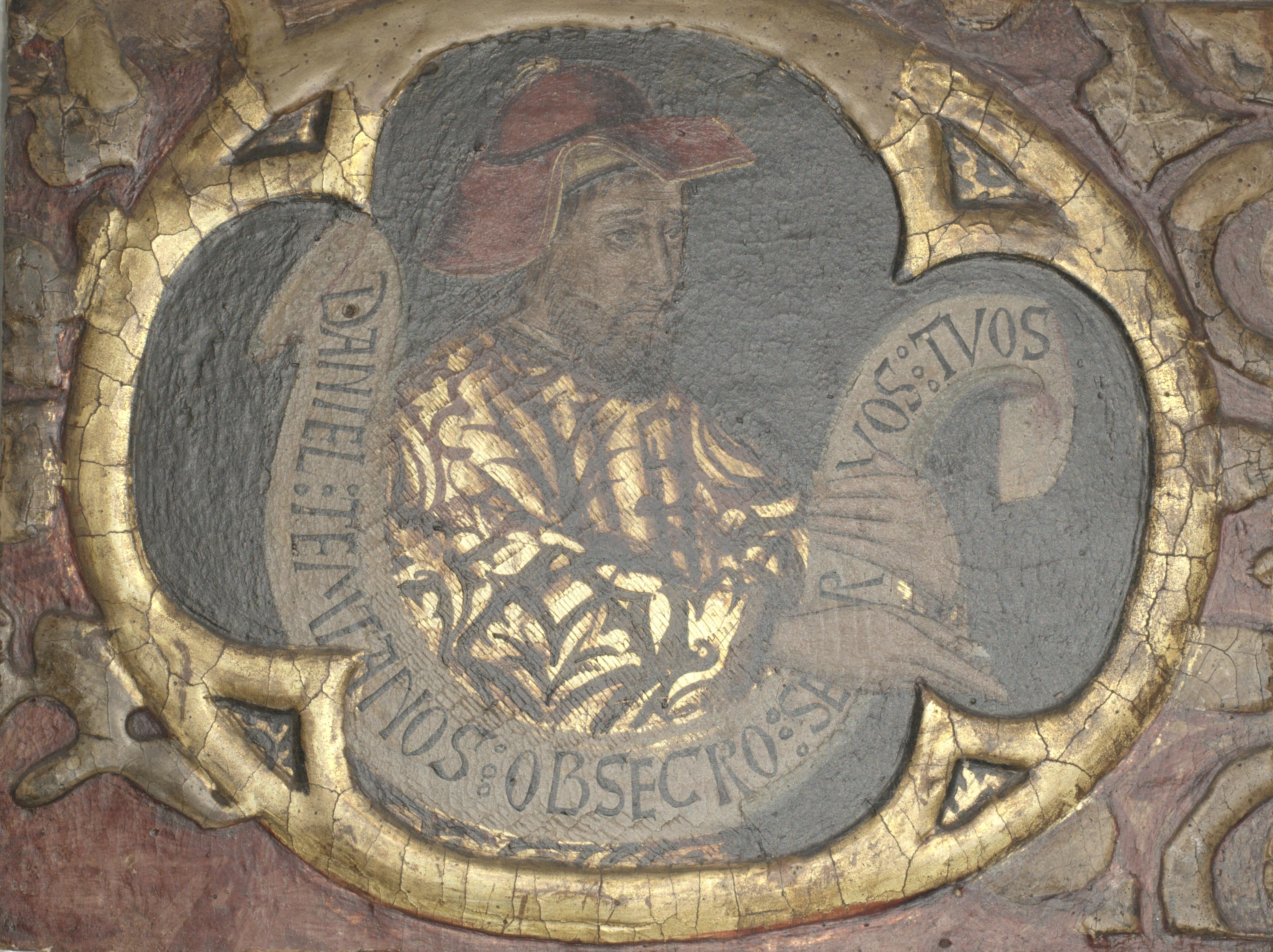}
    \caption{HSH 3rd order}
    \label{fig:retablohl:hsh}
  \end{subfigure}  
  \begin{subfigure}[b]{0.32\linewidth}
\includegraphics[width=\linewidth]{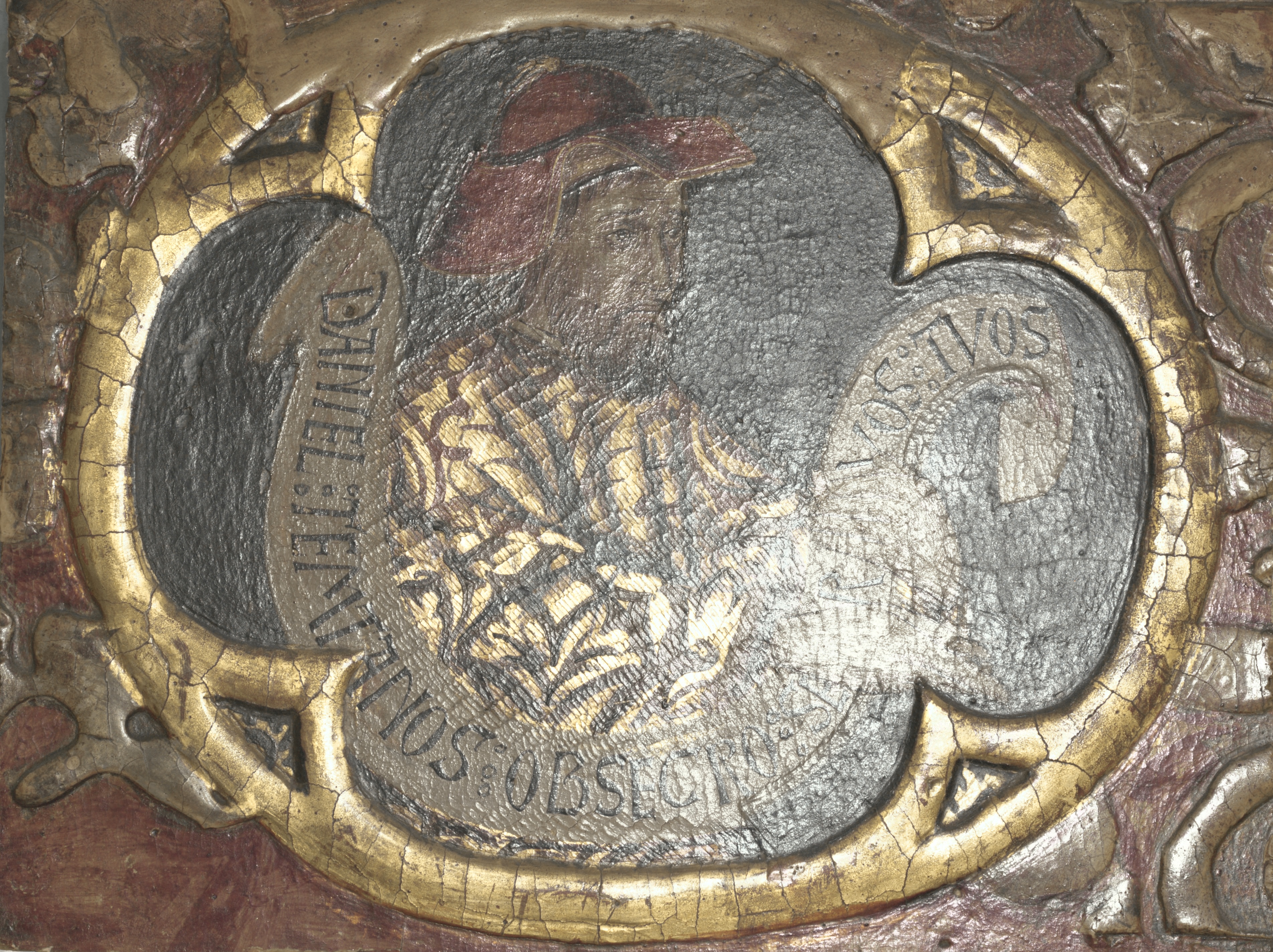}
    \caption{DisK-N\REVISIONADD{eural}RTI-IT (20)}
    \label{fig:retablohl:disk}
  \end{subfigure} 
\begin{subfigure}[b]{0.32\linewidth}
\includegraphics[width=\linewidth]{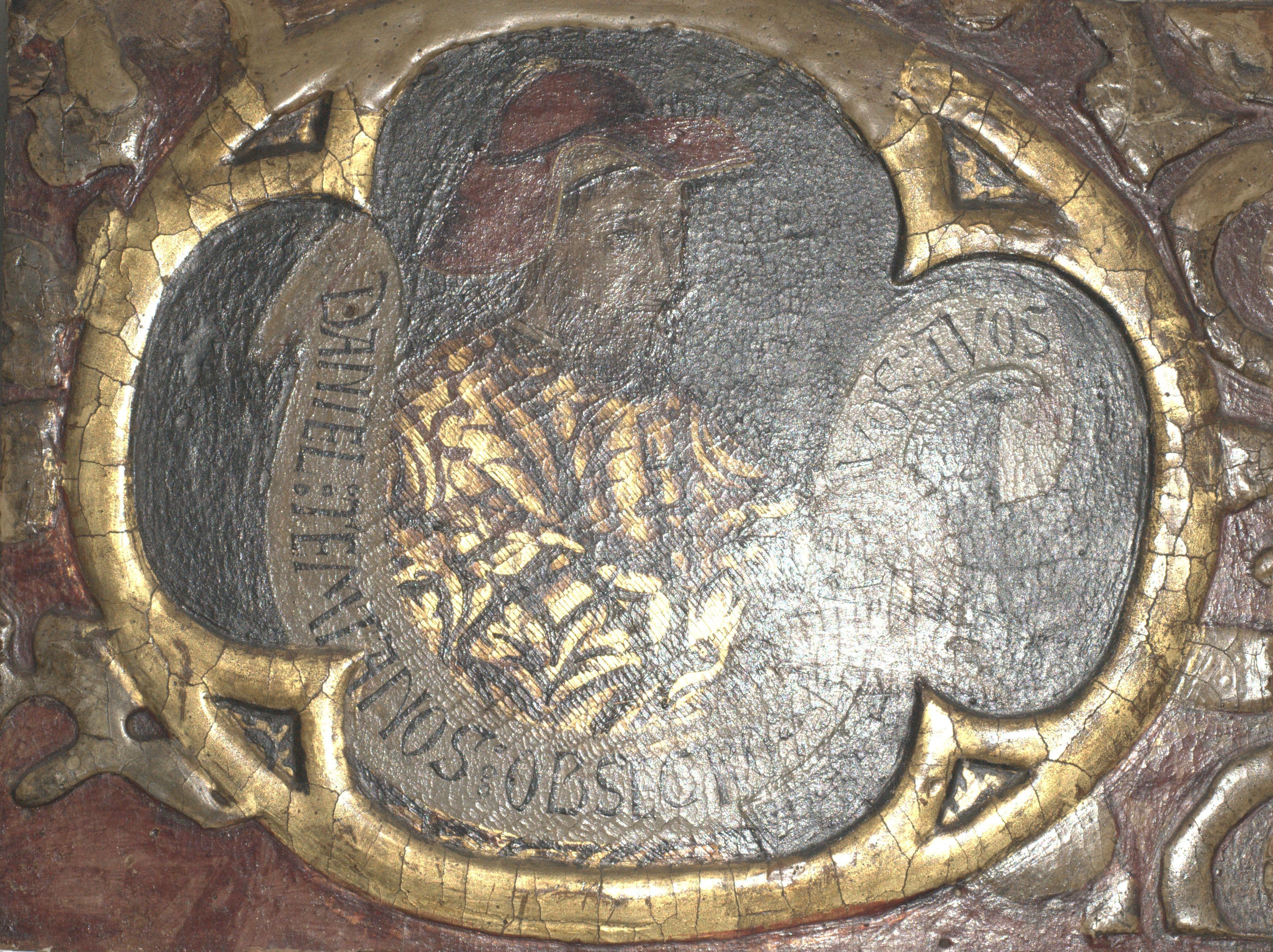}
    \caption{Ground truth}
    \label{fig:retablohl:orig}
  \end{subfigure} 
  \caption{\label{fig:retablohl} Relighting of the Retablo (small) surface with a test light direction not included in the training set. Third-order HSH, despite the use of a heavier per-pixel encoding, fails in representing the correct reflectance behavior (\subref{fig:retablohl:hsh}). The DisK-N\REVISIONADD{eural}RTI-IT result (\subref{fig:retablohl:disk}) is, instead, quite close to the reference image (\subref{fig:retablohl:orig}).}
  
\end{figure*}

The improvements in the relighting quality with the compressed NeuralRTI method \REVISIONREP{are evident}{, also} when compared with the best \REVISIONADD{classic RTI method (\autoref{fig:retablohl})}. The third-order HSH cannot reproduce the highlights and the difference in the reflectance of different materials (a). The same relighting performed with the compressed DisK-NeuralRTI method (b) results in a quite accurate highlight simulation and appears quite similar to the reference image (c). \REVISIONADD{While the more costly NeuralRTI model produces a further, but very slight, quality increase, its complexity cannot ensure full-scale rendering at interactive rates for common viewport sizes (see~\mbox{\autoref{sec:results:chmlic:runtime-performance}}). Thus, DisK-NeuralRTI has the highest quality among the real-time rendering methods.}

\subsubsection{Interactive relighting performances}
\label{sec:results:chmlic:runtime-performance}
NeuralRTI rendering is integrated into the OpenLIME web-based image viewer~\cite{Righetto:2024:EUV}\REVISIONADD{\mbox{\cite{Ponchio:2025:OOF}}}. 
The implementation is based on a custom WebGL 2 shader, running the decoding algorithm in parallel on every pixel on the computer’s graphics card.
The shader loads the decoder's parameters stored after the data-specific training as external variables. The corresponding latent space is a matrix of dimension $H \times W \times K$, where $H \times W$ is the image resolution, and $K$ is the number of features. The shader also loads the matrix elements, previously stored as a set of RGB JPEG images, as samplers. %The shader reads the content of each sampler pixel by pixel. It 
and executes the decoding operations: i.e., scalar product between weights and input vector, sum with the biases, and application of the activation function. 

\begin{figure*}[ht]
  \centering
  \begin{subfigure}[b]{0.49\linewidth}
  \includegraphics[width=\linewidth]{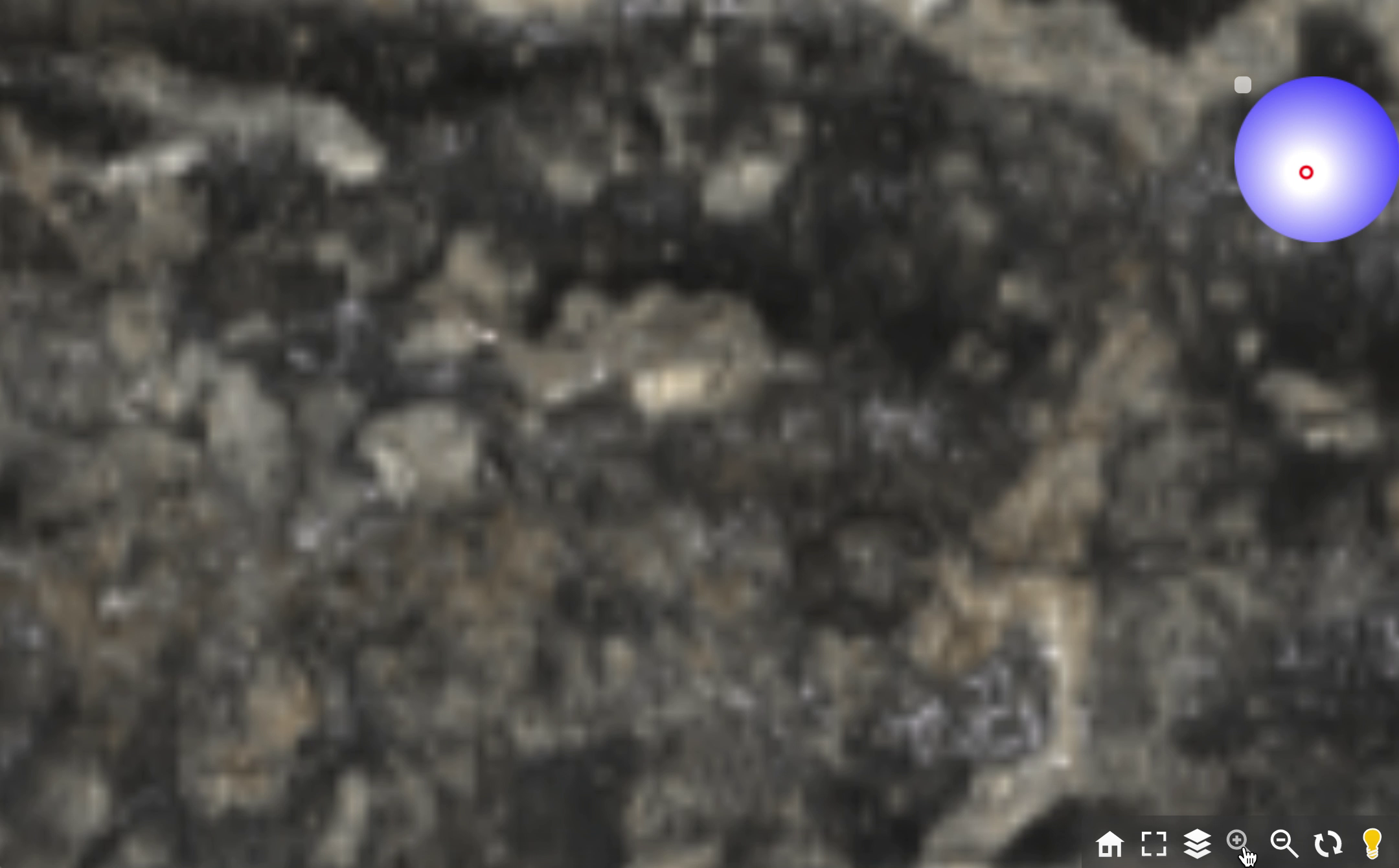}
    \caption{NeuralRTI-50}
    \label{fig:slow}
  \end{subfigure}  
  \begin{subfigure}[b]{0.49\linewidth}
\includegraphics[width=\linewidth]{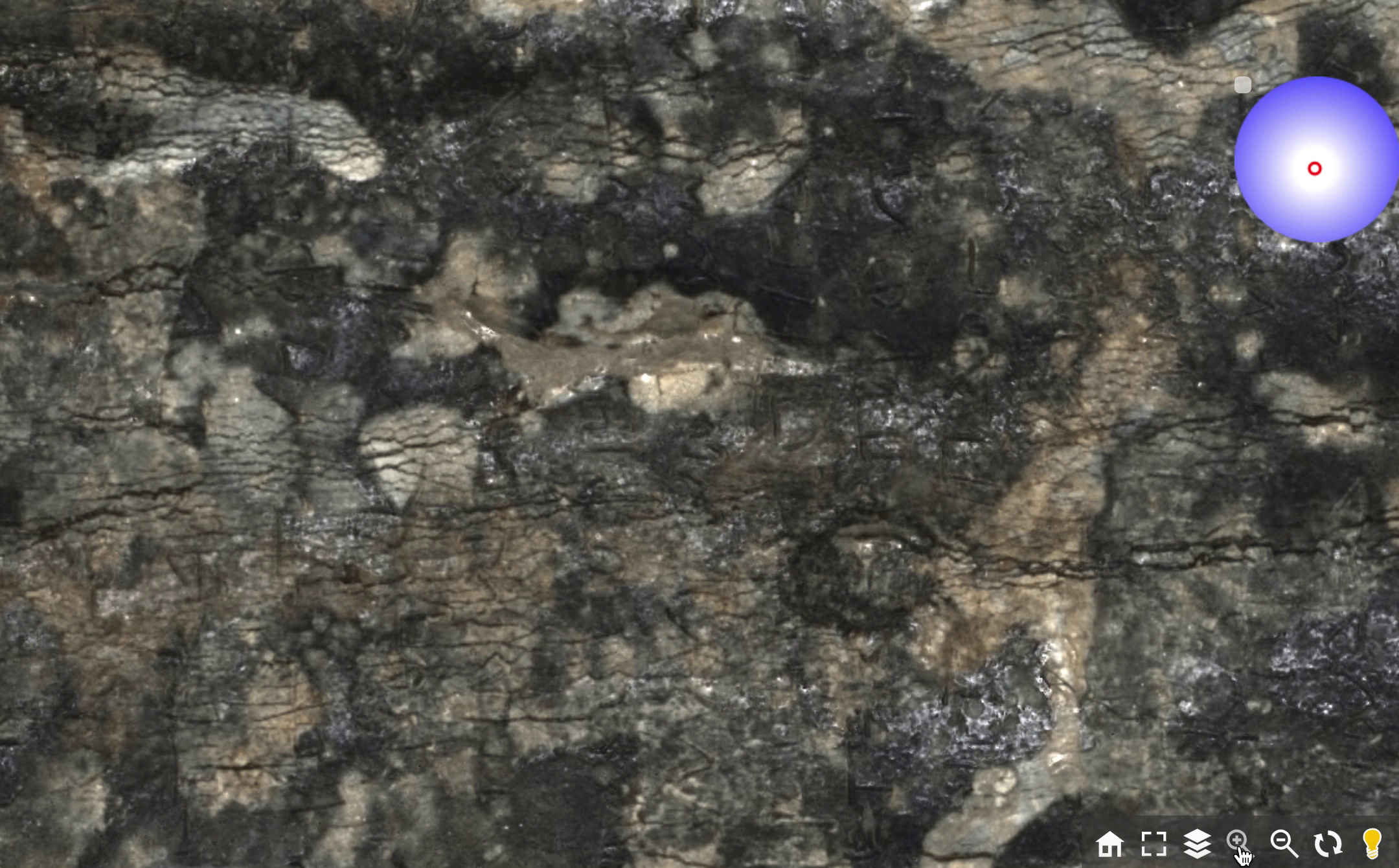}
    \caption{DisK-NeuralRTI (20)}
    \label{fig:fast}
  \end{subfigure} 
  \caption{ Using the adaptive multiresolution rendering of \REVISIONREP{OpenLIME}{OpenLime}, the system dynamically adapts the rendered images' resolution to guarantee interactivity. (a) Snapshot captured in a zooming interaction with the non-compressed NeuralRTI visualization of the Lamina surface. The image is heavily blurred. (b) Snapshot captured in a similar zooming interaction with the compressed version. Images are always sharp. From \cite{drp24}}
  \label{fig:blur}
\end{figure*}

The number of operations required in this procedure is large if the decoder has too many parameters, and it could be unfeasible to use it for large images and large viewports on low-end computers. Particular strategies have been adopted to compensate for thie ~\cite{Righetto:2024:EUV}. The image is split into tiles that are relighted independently, and only the visible tiles on the screen must be rendered. Moreover, the resolutions of the screen of the relighted image are decoupled. When the user interactively modifies the light direction or performs a zoom operation, the application tries to preserve the rendering speed by decreasing the resolution at which the decoding is performed to match a target value (e.g., 20 fps), and, finally, an upscaled version of the relighted image is displayed on the screen. 
When the user stops moving, the full resolution rendering is restored. \autoref{fig:blur} (a) shows this effect on the web viewer: a snapshot captured during a zoom operation is blurred due to the low-resolution decoding and upsampling. Using the DisK-NeuralRTI encoding, there is no need for downsampling, and the snapshot captured during a similar zooming (b) is perfectly sharp.

% --- Tabella valori fps ---
% --------------------------
\begin{table}[ht]
\footnotesize
\setlength{\tabcolsep}{2pt}
\centering
\begin{tabular}{|l|c|c|c|}
\hline
\rowcolor[HTML]{EFEFEF} 
 &
  Lamina  &
  Retablo small &
  Retablo big  \\
  \rowcolor[HTML]{EFEFEF} 
   & ($4328 \times2436$) &
  ($3811 \times 2851$) &
  ($4117 \times 3427$)
  \\ \hline
\cellcolor[HTML]{EFEFEF}
DisK-N\REVISIONADD{eural}RTI (20) & 29.68 & 28.09 & 22.16 \\ \hline
\cellcolor[HTML]{EFEFEF}
NeuralRTI (50) & 1.60 & 1.42 & 1.10 \\ \hline
\end{tabular}%
\caption{Average fps values calculated during relighting of the three high-resolution datasets. From \cite{drp24}.}
\label{tab:fps}
\end{table}

To demonstrate that the NeuralRTI decoder with 20 units per layer and a total of 723 parameters achieves real-time relighting on low-end machines, we performed some tests with the RealRTIHR high-resolution images and the OpenLIME decoder with the image tiling and adaptive resolution options disabled.
We repeated interactive image relighting in sequence with different decoder sizes, collecting the fps values and estimating averages.
The evaluation was done on a MacBook Pro laptop of 2019 (1,4 GHz Intel Core i5 quad-core, graphics card Intel Iris Plus Graphics 645 1536 MB, RAM 8 GB 2133 MHz LPDDR3). The \REVISIONREP{operating}{operative} system was macOS Sonoma version 14.3.1 (23D60), and the web browser was Google Chrome (version 128.0.6613.138).

\autoref{tab:fps} shows the average \REVISIONREP{refresh rate in frames per second}{fps} in the interactive relighting obtained on the RealRTIHR data with the two decoder sizes. Only with the lighter ones, interactive performance is achieved without adaptive resolution.

To evaluate the size of the images that can be relighted on this platform without losing real time performances with no resolution loss we tested the relight of cropped relightable images of different size, approximately increasing the pixel number or one order of magnitude every step, from a \REVISIONREP{$1000 \times 1000$ image (1 million pixels)}{$100 \times 100$ image (10 thousand pixels)} to a $5000 \times 2000$ image (10 million pixels).

\begin{figure}[ht]
\REVISIONIMG{\includegraphics[width=1\linewidth]{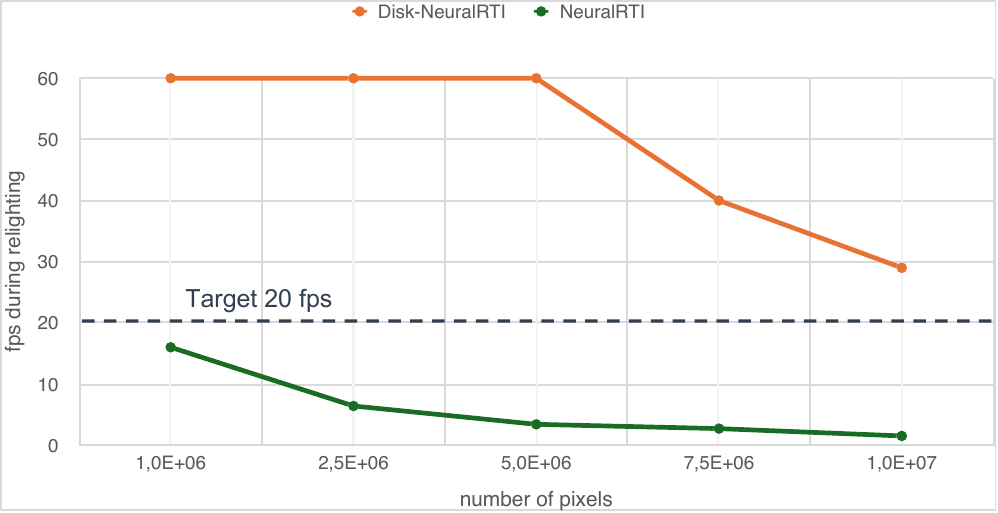}}
\caption{\label{fig:fps_over_pixels} \REVISIONREP{Comparison between speed of exploration with NeuralRTI and DisK-NeuralRTI relighting for different numbers of recomputed pixels. The cost includes all the operations performed by the web viewer, and speed is capped at 60Hz, which is the maximum supported refresh rate in these settings. DisK-NeuralRTI consistency achieves over 30Hz for 1M-10M recomputed pixels per frame, while NeuralRTI is consistently below 20fps, and not interactive starting from 2Mpixels (corresponding to a standard Full HD display.}{Average fps calculation for varying number of pixels simultaneously relighted. On the x-axis is the number of pixels in order of magnitude, i.e., from 10 thousand to 10 million. On the y-axis, the average fps during relighting. From~\mbox{\cite{drp24}}}}
\end{figure}

\REVISIONREP{\mbox{\autoref{fig:fps_over_pixels}} illustrates the performance comparison between NeuralRTI (3303 parameters) and DisK-NeuralRTI (723 parameters) for relighting tasks, measured in frames per second (fps) as a function of the number of pixels recomputed per frame. The evaluation is performed without adaptive resolution reduction~\mbox{\cite{Righetto:2024:EUV}}, and the cost includes the full processing load managed by the web renderer. The maximum achievable frame rate of 60~Hz, set by the display hardware. DisK-NeuralRTI sustains smooth, interactive performance, without any resolution reduction, above 30~fps across a wide range of resolutions (1M to 10M pixels), whereas NeuralRTI remains below 20~fps and drops below interactive thresholds once the pixel count exceeds 2 million, roughly equivalent to Full HD resolution.}{\mbox{\autoref{fig:fps_over_pixels}} shows that the NeuralRTI decoder with  3303 parameters misses the target of 20 fps with 1 megapixel, while the smaller decoder allows an interactive decoding of 10 megapixel arrays without any adaptive solution.}

Video recordings of interactive relighting of RealRTIHR items on the laptop with NeuralRTI(50) and DisK\REVISIONADD{-}NeuralRTI(20) with and without the adaptive resolution tricks can be seen at %\textbf{[Anonymized for review:} URL \url{https://mega.nz/file/wYgynJbQ#Z2P0pmUJWcGayVdOsqfG2FLQpaoCgqI0pfPkKwhYKlo}\textbf{]}
the project link \url{https://tgdulecha.github.io/Disk-NeuralRTI/}.

\subsubsection{Training performance and effect of input image subsampling}

\label{sec:results:chmlic:training-performance}
A possible drawback of increasing the encoder complexity and performing the student training is that the time required to train the final decoder may not be negligible for practical applications. 
We therefore analyzed the time required to complete the training of the teacher and student networks for the full pixel set of large images\REVISIONREP{, as well as}{ and} the effects of downsampling the set of training images, which can speed up the training, on the relight quality.
\autoref{tab:sub} \REVISIONREP{summarizes}{shows} the training times obtained on the 10MP Lamina dataset with the full pixel set and different regular subsampling percentages, obtained on a computer with a 1.4 GHz Intel Core i5 quad-core processor, 32 GB of RAM, and a NVIDIA GeForce RTX 2080 Ti graphics card. 

We can observe that the training times can be large on low-end machines, and a regular subsampling, e.g., 1 pixel every 8x8 tile, can provide a more efficient training, but results in a decrease of the achievable quality, which, however, remains still better than HSH. We plan to investigate smarter pixel sampling strategies and perform further work on training optimization as future work.

\begin{table*}[ht]
\footnotesize
\centering
\begin{tabular}{l|llll|}
\cline{2-5}
 &
  \multicolumn{4}{c|}{\cellcolor[HTML]{EFEFEF}\textbf{Percentage of image pixels}} \\ \cline{2-5} 
 &
  \multicolumn{1}{l|}{\cellcolor[HTML]{EFEFEF}1.56\%} &
  \multicolumn{1}{l|}{\cellcolor[HTML]{EFEFEF}6.25\%} &
  \multicolumn{1}{l|}{\cellcolor[HTML]{EFEFEF}25\%} &
  \cellcolor[HTML]{EFEFEF}100\% \\ \hline
\multicolumn{1}{|l|}{\cellcolor[HTML]{EFEFEF}PSNR/SSIM(Teacher)} &
  \multicolumn{1}{l|}{32.48/0.86} &
  \multicolumn{1}{l|}{33.8/0.87} &
  \multicolumn{1}{l|}{35.21/0.90} &
  37.57/0.92 \\ \hline
\multicolumn{1}{|l|}{\cellcolor[HTML]{EFEFEF}PSNR/SSIM(DisK)} &
  \multicolumn{1}{l|}{34.47/0.88} &
  \multicolumn{1}{l|}{35.7/0.91} &
  \multicolumn{1}{l|}{34.5/0.90} &
  38.33/0.94 \\ \hline
\multicolumn{1}{|l|}{\cellcolor[HTML]{EFEFEF}{\color[HTML]{242424} Teacher Training Time(min.)}} &
  \multicolumn{1}{l|}{6} &
  \multicolumn{1}{l|}{24} &
  \multicolumn{1}{l|}{53} &
  100 \\ \hline
\multicolumn{1}{|l|}{\cellcolor[HTML]{EFEFEF}{\color[HTML]{242424} Student Training Time(min.)}} &
  \multicolumn{1}{l|}{6} &
  \multicolumn{1}{l|}{21} &
  \multicolumn{1}{l|}{83} &
  132 \\ \hline
\rowcolor[HTML]{EFEFEF} 
\multicolumn{1}{|l|}{\cellcolor[HTML]{EFEFEF}{\color[HTML]{242424} \textbf{Total Training Time(min.)}}} &
  \multicolumn{1}{l|}{\cellcolor[HTML]{EFEFEF}12} &
  \multicolumn{1}{l|}{\cellcolor[HTML]{EFEFEF}45} &
  \multicolumn{1}{l|}{\cellcolor[HTML]{EFEFEF}136} &
  232 \\ \hline
\end{tabular}
\caption{\label{tab:sub} Training times of the DisK-N\REVISIONADD{eural}RTI model with the improved teacher with the full pixel set and regularly subsampled data at different ratios.}
\end{table*}

%\FIXME{LEONARDO/TINSAE: potete mettere link anonimo a video eventualmente}

\section{Discussion}
\label{sec:discussion}
The original NeuralRTI has shown its capability to preserve the compression rate of previous classic RTI solutions with a much improved reproduction of real reflectance properties of surfaces, especially high-frequency ones~\cite{Dulecha:2020:NRT}.
However, the relatively costly custom decoder used by the technique to perform the relighted image rendering may create an annoying latency for high-resolution images on low-end devices. Previous work aimed at integrating the method in an online viewer solved the issue by adapting the resolution of the rendered window to the desired frame rate during the interaction~\cite{Righetto:2024:EUV}, at the cost of detail loss during light or camera movements, especially on large screen displays driven by commodity graphics boards. This situation is very common, for instance, when experts analyze models on laptop/mobile devices or when relightable image viewers are used for museum exploration on nowadays 4K, or even 8K, touch screens. 

Using the proposed network compression approach based on Knowledge Distillation, we showed how to strongly reduce the decoding time, making it possible to render large images in real time with interactive performance on standard PCs without lowering the resolution.

Previous works have demonstrated that regressing the reflectance behavior by directly training a small network on raw reflectance data provides suboptimal results, due to the difficulty of the error landscape. Our work is the first attempt to apply automated network compression approaches to  NeuralRTI, and it has given promising results.

Our tests show that the compressed encoding can guarantee smooth interactive relighting with a higher resolution than the one displayed on 4K UHD screens using low-end hardware. This makes it practical for professional cultural heritage and engineering applications.

\REVISIONADD{DisK-NeuralRTI encodings can be generated from the same input data used by traditional RTI approaches, such as PTM and HSH. This compatibility allows DisK-NeuralRTI to serve as a drop-in replacement for these methods in relighting frameworks, offering significantly higher visual quality while maintaining comparable storage, transmission, and rendering performance
}

\REVISIONADD{
While the approach proposed in this work works well and the results obtained are promising, it is also useful to point out the limitations of our work and show directions for future improvements. A first one is related to the choice of the teacher network.}
%However, it is certainly possible to improve it. 
Our initial experiments~\cite{drp24} used the same original NeuralRTI model as a teacher network. This architecture was initially designed with a light encoder and decoder architecture to achieve acceptable training times and interactive relighting. In this work, we have shown that by applying knowledge distillation from a deeper teacher architecture, it is possible to improve the quality of the results further or make the decoding even more efficient. Future work may improve in this area by also evaluating further modifications, both in the entire teacher network and in the encoder size of the student network, which is not used at run-time.

\REVISIONADD{A second potential issue is related to the training time.}
Like other learning-based methods, NeuralRTI takes longer to generate a representation compared to traditional fitting-based approaches such as PTM or HSH, as it must optimize the many parameters of a non-linear function through loss minimization computed on input data. With distillation, the cost is increased, since we need to first optimize the teacher network to later optimize the desired student network. Using a more complex teacher further increases this cost. 
However, this is not a major concern for end-user applications, since training is done only once, is typically not time-critical, as opposed to exploration, and is still faster than acquiring a complex reflectance field of an object, especially when using GPU-accelerated nodes. 
In this article, we have shown that it is possible to reduce learning times by restricting training to a subset of the pixels, exploiting the redundancy present in the images. The selected method, which just performs data-independent subsampling, achieves performances higher than non-neural competitors, but decreases the quality relative to the best results achievable when training with all pixels.

\REVISIONADD{
To address this, future work will focus on optimizing the selection of training pixels based on their information content. Similar strategies have shown promising results in learning-based compression techniques for volume rendering~\mbox{\cite{Gobbetti:2012:CCO,Diaz:2020:ISE}}.}
\section{Conclusion}
\label{sec:conclusions}

We have shown \REVISIONREP{how}{the potential of using} knowledge distillation \REVISIONREP{can}{to} create more efficient Neural Reflectance Transformation Imaging (RTI) decoders for interactive object exploration in cultural heritage applications. While neural representations have \REVISIONREP{demonstrated in the past}{shown} superior image quality at \REVISIONREP{storage costs comparable}{comparable storage costs} to traditional models like PTM or HSH, their decoding costs have often hindered their practical usage for real-time high-resolution exploration of large models on high-pixel-count displays. In contrast to previous manual attempts to tune network size, our approach leverages a knowledge distillation framework, where a smaller student network is trained to mimic the output of a larger, more complex teacher network, resulting in a compressed model that retains high-quality relighting capabilities.

The adoption of neural distillation has been shown to overcome the limitations of manually tuning the decoding network complexity, a process that often involves trade-offs between quality and efficiency. \REVISIONDEL{With distillation, a student network with a compact and fast decoder is guided to learn from a high-capacity teacher network, effectively preserving relighting quality while significantly reducing inference cost. The resulting decoder is more suitable for practical deployment in real-world cultural heritage applications, where high-quality rendering and performance efficiency are both essential to enable viewers to perceive subtle illumination changes during light or camera manipulations.}

To evaluate our optimization strategy, we extended the evaluation framework used in our previous work~\cite{drp24}\REVISIONADD{, also} by incorporating \REVISIONREP{four}{two} new datasets, introducing a comprehensive benchmark dubbed \emph{RealRTIHR}. This benchmark
covers a diverse range of surface types and material properties, and is designed to assess both relighting quality and computational efficiency under realistic usage conditions. \REVISIONDEL{Our experimental results show that the student networks retain a high degree of fidelity to the teacher’s output, even when trained on a reduced subset of image pixels, lowering data preparation costs and further enhancing the practicality of the approach on large image sets.} The resulting compact model, when integrated in interactive web-based viewers, allows it to explore large, high-resolution relightable images interactively on standard hardware while preserving a high relighting quality during interaction. This contribution represents a significant step toward making neural relightable image representations more accessible and deployable in real-world cultural heritage contexts, where rendering performance, storage efficiency, and visual accuracy are all critical. 

%===========================================================================
% Acks
%===========================================================================

%\begin{spacing}{0.8}
%\begin{footnotesize}

\section*{Acknowledgments}
\ANONYMIZE{This study was partially funded by the consortium iNEST funded by the EU Next-GenerationEU PNNR M4C2 Inv1.5 – D.D.~1058 23/06/2022, ECS00000043 and by the project REFLEX (PRIN2022) funded by the EU Next-GenerationEU PNRR M4C2 Inv. 1.1. EG and RP also acknowledge the contribution of Sardinian Regional Authorities under project XDATA (RAS Art9 LR 20/2015). We thank Prof. Attilio Mastrocinque and the Civic Archaeological Museum of Milan for access to the lead sheet, the National Archaeological Museum of Cagliari for access to the retablos, Tomasz Łojewski (AGH University of Science and Technology, Kraków) for the Oseberg find artifact, and Opificio Delle Pietre Dure for the bronze panel. We thank Fabio Bettio (CRS4), Fabio Marton (CRS4), and Federico Ponchio (ISTI-CNR) for their work in developing \REVISIONREP{OpenLIME}{OpenLime} and for the integration of Neural RTI rendering within the framework.
We acknowledge ISCRA for awarding this project access to the LEONARDO supercomputer, owned by the EuroHPC Joint Undertaking, hosted by CINECA (Italy).}
%We acknowledge the CINECA award under the ISCRA initiative (BERET project) for the availability of high-performance computing resources and support.

%\end{footnotesize}
%\end{spacing}

%%%%%%%%%%%%%%%%%%%%%%%%%%%%%%%%%%%%%%%%%%%%%%%
\bibliographystyle{cag-num-names}
\bibliography{cag2025-mainpaper.bib}

\end{document}